%% file: main.tex
\documentclass{article}

\PassOptionsToPackage{numbers, compress}{natbib}

\usepackage[preprint]{neurips_2025}

\usepackage[utf8]{inputenc} 
\usepackage[T1]{fontenc}    
\usepackage{hyperref}       
\usepackage{url}            
\usepackage{booktabs}       
\usepackage{amsfonts}       
\usepackage{nicefrac}       
\usepackage{microtype}      
\usepackage{xcolor}         
\definecolor{custompurple}{HTML}{B5739D}

\definecolor{customgreen}{HTML}{67AB9F}
\usepackage{multirow}
\usepackage{graphicx}
\usepackage{amsmath}
\usepackage{amsthm}
\newtheorem{proposition}{Proposition}
\usepackage{algorithm}
\usepackage{algorithmic}
\usepackage{bibunits}

\title{CausalPlan: Empowering Efficient LLM Multi-Agent Collaboration Through Causality-Driven Planning }

\author{%
  Minh Hoang Nguyen\thanks{Corresponding author.},
  Van Dai Do,
  Dung Nguyen,
  Thin Nguyen,
  Hung Le
  \\
  Applied Artificial Intelligence Initiative (A$^2$I$^2$) \\
  Deakin University \\
  Geelong, Australia \\
  \texttt{\{s223669184, s223540177, dung.nguyen, thin.nguyen, thai.le\}@deakin.edu.au} \\
}

\begin{document}

\begin{bibunit}[plainnat]

\maketitle

\begin{abstract}
Large language model (LLM) agents—especially smaller, open-source models—often produce causally invalid actions in collaborative tasks due to their reliance on surface-level correlations rather than grounded causal reasoning. This limitation undermines their performance in terms of coordination and planning in dynamic environments. We address this challenge with CausalPlan, a two-phase framework that integrates explicit structural causal reasoning into the LLM planning process. At the core of CausalPlan is the Structural Causal Action (SCA) model, which learns a causal graph from agent trajectories to capture how prior actions and current environment states influence future decisions. This structure is then used to guide action selection by assigning causal scores to LLM-generated proposals, reweighting them accordingly, or falling back to causally grounded alternatives when needed. By embedding this causal knowledge directly into the decision loop, CausalPlan constrains planning to intervention-consistent behaviours without requiring fine-tuning of the LLM itself. We evaluate CausalPlan on the Overcooked-AI benchmark across five multi-agent coordination tasks and four LLMs of varying sizes: Gemma-7B, Llama-8B, Qwen-14B, and Llama-70B. Experimental results show that CausalPlan consistently reduces invalid actions and improves collaboration in both AI-AI and human-AI settings, outperforming strong reinforcement learning baselines. Our findings highlight the value of causality-driven planning for deploying efficient, interpretable, and generalisable multi-agent LLM systems.
\end{abstract}

\section{Introduction}
\label{sect: intro}
\input{input/intro}

\section{Related Works}
\label{sect: related work}
\input{input/related_work}


\section{Method}
\label{sect: Method}

\subsection{Preliminaries}
\input{input/preliminaries}

\input{input/method}

\section{Experiments}
\label{sect: experiments}
\input{input/experiments}

\section{Conclusion, Limitations, and Impacts Discussion}
\label{sect: conclusion}
\input{input/conclusion}

\putbib[main]  

\end{bibunit}

\newpage

\newpage
\appendix
\begin{bibunit}[plainnat]

\begin{center}
    \Large\bfseries Appendix for \\
    \textit{CausalPlan: Empowering Efficient LLM Multi-Agent Collaboration Through Causality-Driven Planning}
\end{center}

\renewcommand{\thetable}{A\arabic{table}}  
\renewcommand{\thefigure}{A\arabic{figure}}  
\setcounter{table}{0}  
\setcounter{figure}{0}  

\input{input/Appendix}

\clearpage
\pagebreak

\putbib[appendix]  

\end{bibunit}

\end{document}

%% file: input/intro.tex
Large Language Models (LLMs) have demonstrated significant success across various natural language processing tasks~\citep{openai2024gpt4technicalreport, zhao2023survey, deepseekai2025deepseekr1incentivizingreasoningcapability}. Recently, there has been growing interest in leveraging LLMs as decision-makers, particularly within multi-agent frameworks for executing interactive planning tasks. Research in this area investigates how multiple LLM agents can collaborate and make decisions, with notable works including the emergence of human-like behavior~\citep{park2023generative}, integrated pipelines for cooperative tasks~\citep{zhang2023building}, graph-based coordination~\citep{qian2024scaling}, and human-AI collaboration frameworks~\citep{zhang2024proagent}.



A major challenge in multi-agent learning is promoting effective collaboration (creating a generalized agent that can collaborate with a wide range of unseen partners)~\citep{legg2007universal, hu2020other}. The ultimate objective is to develop an agent that can seamlessly interact and cooperate with humans. Two of the most prevalent approaches to address this are self-play (SP)~\citep{tesauro1994td} and population-based training (PBT)~\citep{jaderberg2017population, carroll2019utility, strouse2021collaborating, zhao2023maximum}, with some approaches combining both techniques~\citep{li2023cooperative}. While effective in training environments, these classic (non-LLMs reinforcement learning (RL)) multi-agent coordination approaches often struggle to generalize to cases of unseen collaborators. This limitation stems from the fact that agents are typically overfitted to the specific behaviors and distributions of their training partners, reducing robustness in open-ended environments~\citep{zhang2024proagent}. Moreover, these methods also neglect domain knowledge that could accelerate learning and sample efficiency that could support faster adaptation and better collaboration.

\begin{figure}[tbp]
\begin{center}
    \includegraphics[width=0.8\textwidth, height=3cm]{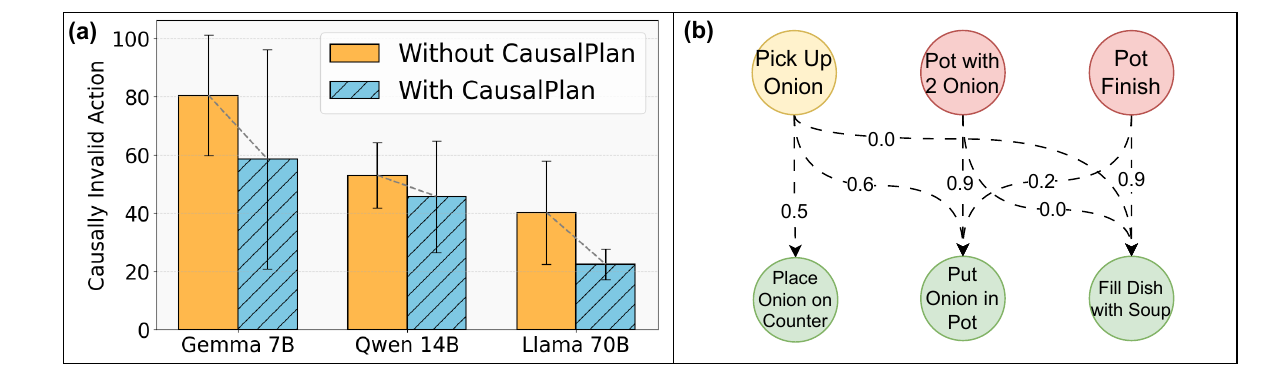}  
\end{center}
\caption{
\textbf{(a)} Evaluation on the Cramped Room layout showing how the number of causally invalid actions changes with LLM size, averaged over four seeds. The LLM agent plays as Player~1, while Player~0 is controlled by COLE~\citep{li2023cooperative}. CausalPlan significantly reduces the number of invalid moves. 
\textbf{(b)} Simplified causal graph discovered by CausalPlan for the same layout. Yellow and red nodes indicate parent actions and states, respectively, while green nodes denote child actions. ``Pick Up Onion'' strongly influences ``Put Onion in Pot'' (0.6) and ``Place Onion on Counter'' (0.5), but not ``Fill Dish with Soup'' (0). The state ``Pot with 2 Onions'' strongly drives ``Put Onion in Pot'' (0.9), while ``Pot Finished'' weakly affects it (0.2) but strongly influences ``Fill Dish with Soup'' (0.9).
}
\label{fig:1_counting_failure}  
\end{figure}

Large Language Model (LLM)-based agents, trained on vast and diverse datasets containing rich common-sense and procedural knowledge, have recently emerged as promising tools for multi-agent planning. Compared to traditional multi-agent learning methods—which often struggle with generalization, sample inefficiency, and lack of transferable priors~\citep{zhang2024proagent}—LLMs, especially powerful closed-source ones like GPT-4~\citep{zhang2024proagent}, demonstrate impressive performance in collaborative environments such as Overcooked-AI~\citep{carroll2019utility}. However, despite these strengths, a persistent limitation remains: LLM agents often lack robust causal reasoning~\citep{gao2023chatgpt}. This shortcoming leads to causally invalid plans that violate physical constraints or temporal dependencies of the environment—for example, placing an onion into a pot before picking it up, or executing an action absent from the task instructions. Even large closed-source models like GPT-4 occasionally exhibit such errors, but the problem is significantly more pronounced in smaller, open-source models due to their limited capacity and narrower training coverage. Crucially, smaller and open-source LLMs remain highly attractive for enterprise and resource-constrained settings because of their accessibility, lower deployment costs, and greater controllability. Yet, their higher incidence of causally invalid actions often undermines their collaborative performance. As illustrated in Figure~\ref{fig:1_counting_failure}(a), our evaluation of multiple open-source LLMs with varying parameter sizes in the Cramped Room task of the Overcooked-AI testing suite~\citep{carroll2019utility} reveals that even large open-source models like Llama-70B still produce a substantial number of causally invalid actions. These findings underscore a clear gap: to fully unlock the potential of LLMs—particularly smaller and more deployable ones—in collaborative multi-agent planning, their causal reasoning capabilities must be significantly enhanced.

Unlike LLMs, humans draw on intergenerational causal knowledge to inform better decision making - in particular, our ability to plan and choose the next action is deeply grounded in this inherited understanding of cause and effect, shaped over generations. From the perspective of causality, causal knowledge refers to understanding how different variables - such as states and the actions of an agent - influence each other ~\citep{pearl2009causality}. By modeling these dependencies explicitly, agents can anticipate the effects of their own and others’ decisions, avoid redundant actions, and better coordinate toward a shared goal. These causal dependencies are naturally represented by a graph-like structure known as a causal graph $\mathcal{G}$, which captures the directional relationships between variables—where nodes denote elements such as actions or states, and edges represent causal influences—and its corresponding Structural Causal Model (SCM),  a formal framework that defines how each variable is generated from its parent variables in the graph~\citep{pearl2009causality} (refer to Figure~\ref{fig:1_counting_failure} (b) for an example of the causal graph discovered by our method). Together, the graph and structural equations enable agents to reason about interventions and understand the causal dynamics of the system. An SCM can be identified through causal discovery, which learns both the graph structure—captured by structural parameters—and the underlying data-generating mechanisms—captured by functional parameters—from observational data~\citep{pearl2009causality, ke2019learning, peng2022causality}. In contrast, causal inference leverages this learned causal structure to support downstream decision-making tasks~\citep{pearl2009causality}. Building on this foundation, we investigate multi-agent collaboration with LLM-based agents and pioneer a novel approach that leverages causal knowledge to enhance planning. We hypothesize that better causality-driven planning will enable more effective collaborative cooperation. Ultimately, we want to answer the question of: \textit{''How can we equip LLMs with causal knowledge to enable effective multi-agent collaboration?''}. 

\textbf{Overview of CausalPlan.} To answer the question, we aim to leverage the dependencies between \emph{both} agents’ states and past actions to inform future planning in a collaborative setting. For instance, before serving a plate of soup (next action), one must first fill the dish with soup (the previous action) and hold the filled dish (current state). In a multi-agent context, if the partner agent is already holding a filled dish (current state of the other agent), our agent may instead focus on complementary tasks rather than duplicating effort. Specifically, our causal graph includes the partner agent’s state (Fig.~2), enabling reasoning over both agents’ states and actions to coordinate effectively. To model such causal dependencies, we introduce the Structural Causal Action (SCA) model—an extension of SCM tailored to capture the causal relationships between previous actions, current states (for both agents), and future actions. Building on this foundation, we propose CausalPlan, a framework comprising two key phases inspired by the processes of causal discovery and inference: Causal Action Structure Learning and Agent Planning with Causal Knowledge. In the first phase, we learn the SCA model by iteratively optimizing both its generating and structural parameters, drawing on established techniques for SCM discovery. Once trained, SCA yields a \emph{Causal Action Matrix} $\mathcal{M}$ that encodes how past actions and current observations of \emph{both agents} causally influence each possible next action, in the form of causal scores. In the second phase, we integrate this causal knowledge into planning: at each step, we query $\mathcal{M}$ using the current state and past actions of both our agent and the partner agent. We address the possibility of causally invalid actions in one of two ways. \emph{Causal-Aware Action Planning} adjusts the LLM’s raw action probabilities by re-weighting them with the causal scores from the matrix, then resamples to choose the next action that is both valid and complementary to the partner’s role. If there are no valid actions, we fall back to \emph{Causal Backup Action}, which selects the action with the highest causal probability. An overview of this two-phase framework is shown in Fig.~\ref{fig:2_framework}.

We evaluate CausalPlan on the Overcooked-AI benchmark~\citep{carroll2019utility} using four open-source LLMs—Gemma-7B, Llama-8B, Qwen-14B, and Llama-70B—across both AI-AI and human-AI collaboration settings. Empirical results show that CausalPlan consistently improves planning performance and reduces invalid actions, even for the smallest LLMs without finetuning or training multi-agents using traditional RL approaches. Our main contributions are:
\begin{itemize}
    \item We identify a core failure mode of LLM agents in multi-agent collaboration—generating causally invalid plans—and propose causal reasoning as a principled remedy.
    \item We introduce \textbf{CausalPlan}, a two-phase framework that integrates causal discovery and inference to enhance LLM agent planning and collaboration.
    \item We demonstrate, through extensive experiments, that CausalPlan improves performance across multiple model sizes and collaboration scenarios, outperforming strong RL baselines.
\end{itemize}

%% file: input/related_work.tex
\textbf{Reasoning and planning with LLM agents.}
The rise of LLMs has enabled applications in both single and multi-agent settings~\citep{xi2025rise}. A core focus of recent work lies in improving reasoning and action planning capabilities, which are also central objectives of this paper. The works in a single agent setting focus on improving reasoning through chain of thought prompting~\citep{wei2022chain, kojima2022large}, self-consistency~\citep{wang2022self}, and problem decomposition~\citep{zhou2022least}. LLMs have also been applied to robotic planning~\citep{ahn2022can}, integrated reasoning and acting, and reflection-based learning~\citep{shinn2023reflexion}. ~\citet{zhu2024knowagent} and \citet{qiao2024agent} leverage memory of past actions and states to improve planning, though these prior works focus on single-agent scenarios. In contrast, our work targets multi-agent environments. In multi-LLM-agent research,~\citet{park2023generative} pioneered human-like interactions where agents remember and act autonomously.~\citet{zhang2023building} proposed a fully automated cooperative framework with perception, communication, and planning, while~\citet{cross2024hypothetical} incorporated Theory of Mind (ToM).~\citet{qian2024scaling} explores scaling up using different collaboration topologies.~\citet{zhang2023building} is one of the latest works that examines human-LLM cooperation. Although we address a similar challenge, our work pioneers the incorporation of causal knowledge in multi-LLM-agent collaboration.

\textbf{Zero-shot multi-agent coordination.}
Zero-shot multi-agent coordination aims to train agents that can collaborate with unseen partners, human or AI. A classic method is Self-Play (SP)~\citep{tesauro1994td, carroll2019utility}, where agents train by interacting with themselves. While effective, SP often fails to generalize to unfamiliar partners. Population-Based Training (PBT)~\citep{jaderberg2017population} promotes learning by diversifying the population of training agents, but it still suffers from overfitting to seen partners. Recent methods combine SP and PBT to boost diversity, such as Fictitious Co-Play (FCP)~\citep{strouse2021collaborating} and Maximum Entropy Population (MEP)~\citep{zhao2023maximum}. COLE~\citep{li2023cooperative} shifts focus to strategic policy selection during training. However, these methods are generally computationally expensive and lack interpretability. ~\citet{zhang2024proagent} shows that LLM-based agents can excel in zero-shot tasks by leveraging rich language knowledge. While this demonstrates the potential of language-based agents, LLMs tend to lack the ability for causal reasoning~\citep{gao2023s3}. To address this challenge, we propose a causal learning approach that enhances coordination for open-sourced LLMs, aiming to balance performance, efficiency, and interpretability.

\textbf{Causality in decision making.}
Causal reasoning has gained attention for improving AI decision-making in both single and multi-agent settings. In single-agent domains, counterfactual-based methods~\citep{pitis2020counterfactual,pitis2022mocoda} were used for data augmentation. ~\citet{corcoll2020disentangling} use causality to construct variable hierarchies,~\citet{zhang2023interpretable} redistributes rewards based on causal impact, and~\citet{seitzer2021causal, zhang2021deep} incorporate causal signals into reward shaping.~\citet{peng2022causality} learns causal graphs to define hierarchical subgoals, improving efficiency. In multi-agent settings, social influence is used as a proxy for inter-agent causality, promoting cooperation~\citep{jaques2019social}. Subsequent works use action influence and reward redistribution to encourage coordinated behaviors~\citep{du2024situation, zhang2024causality}. Although effective, these methods are based on traditional RL frameworks. In contrast, we try to integrate causal reasoning into LLM-based agents, requiring a different method.

%% file: input/preliminaries.tex
\textbf{Markov Decision Process.} A two-player Markov Decision Process (MDP) is defined as \( (\mathcal{S}, \{ \mathcal{A}^i \}, P, \gamma, R) \), where \( \mathcal{S} \) is the state space, \( \mathcal{A}^i \) is the action set for agent \( i \in \{1,2\} \), \( P \) defines the transition dynamics, \( \gamma \in [0,1) \) is the discount factor, and \( R: \mathcal{S} \times \mathcal{A} \) is the reward function where \(\mathcal{A}=\mathcal{A}_1 \times \mathcal{A}_2\) is the joint action space. We assume a factored state space \( \mathcal{S} = \mathcal{S}^{\text{agent}} \times \mathcal{S}^{\text{env}} \), where \( \mathcal{S}^{\text{agent}} \) is the state of the agent (both agent 1 and 2) and \( \mathcal{S}^{\text{env}} \) the state of the environment. Let \( S = |\mathcal{S}| \) and \( A = |\mathcal{A}| \) denote the dimensions of $\mathcal{S}$ and $\mathcal{A}$, respectively. At each timestep \( t \), each agent \( i \in \{1,2\} \) observes the current state \( s_t = (s^{\text{agent}}_t, s^{\text{env}}_t) \) and selects an action according to its policy \( \pi^i(a^i_t \mid s_t) \), forming the joint action \( a_t = (a^1_t, a^2_t) \). A trajectory is given by \( \tau = (s_1, a_1, s_2, a_2, \dots) \), and the objective is to maximize the expected cumulative reward, \( \mathbb{E} \left[ \sum_t R(s_t, a_t) \right] \).

\textbf{Causality and Structural Causal Model.}
Causality studies the relationships between variables and events~\citep{pearl2009causality}. The Structural Causal Model (SCM) framework represents causal relationships in a system, where for a set of variables \( V = \{V_1, \dots, V_M\} \), each variable \( V_i \) is defined as \( V_i := f_i(\text{Pa}_{\mathcal{G}}(V_i), \varepsilon_i) \), with \( \{f_1, f_2, \dots, f_M\} \) being generating functions, \( \text{Pa}_{\mathcal{G}}(V_i) \) the parents of \( V_i \) in the causal graph \( \mathcal{G} \), and \( \{\varepsilon_1, \dots, \varepsilon_M\} \) noise terms~\citep{pearl2009causality}. The directed acyclic graph (DAG) causal \( \mathcal{G} = \{V, E\} \) contains edges \( e_{ji} \in E \), where \( e_{ji} = 1 \) indicates that \( V_j \) causes \( V_i \), and \( e_{ji} = 0 \) otherwise~\citep{pearl2009causality}. SCMs are often learned from data by modeling the generating functions \( f_i \) as neural networks parameterized by generating parameters \( \delta \)~\citep{ke2019learning, peng2022causality, zhang2023interpretable}, with causal edges \( e_{ji} \) inferred if the binary adjacency indicator \( \eta_{ji} > \gamma \), where \( \gamma \) is a confidence threshold~\citep{ke2019learning, peng2022causality, zhang2023interpretable}.


%% file: input/method.tex
\subsection{CausalPlan Framework: Structure and Phases}
The CausalPlan framework comprises two main phases.  
In Phase 1, we construct the SCA Model and derive from it the Causal Action Matrix~$\mathcal{M}$.  
In Phase 2, at each timestep~$t$, we first provide the current observation~$s_t$ to the LLM agent and prompt it to analyze the observation.  
Both the observation~$s_t$ and the resulting analysis are then used as inputs for a second prompt, where the agent is asked to generate a set of actions  
(details of this two-prompt process are described in Appx.~\ref{sect A "1P" vs "2P"}).  
We, then, leverage~$\mathcal{M}$ to modify the agent's action selection, either through reweighting LLM-generated proposals using causal scores in the Causal-Aware Action Planning module, or by reverting to causally grounded alternatives in the Causal Backup Action module (see Appx.~\ref{sect A CausalPlan Details} for the full algorithms).

\begin{figure}[t!]
\centering
\includegraphics[width=1.02\textwidth, height=6.5cm, trim={50 0pt 0pt 0pt}, clip]{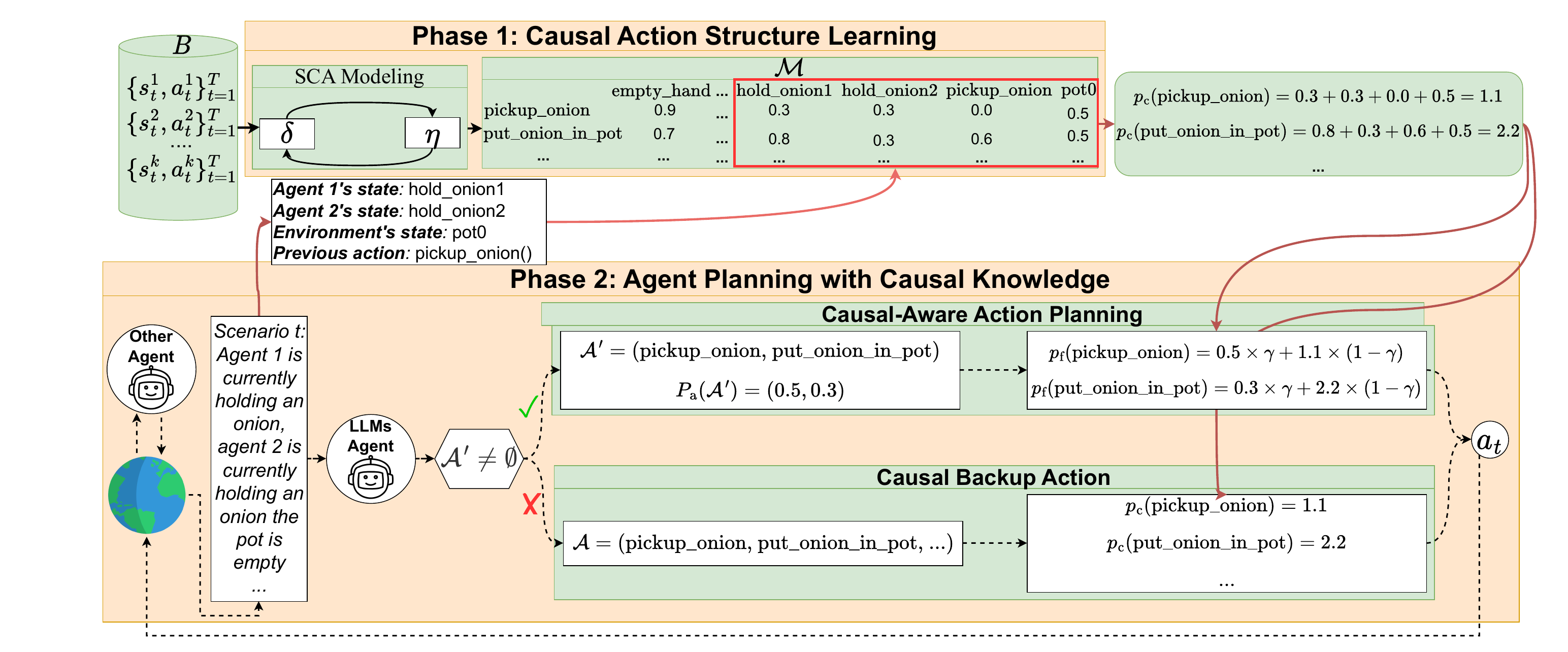}  
\caption{
\textbf{Overview of the CausalPlan Framework}. The process begins with a dataset \( B \) collected by a behavior policy \( \pi_{\beta} \). In Phase 1 (\textit{Causal Action Structure Learning}), we train the Structural Causal Action (SCA) Model by optimizing generating (\( \delta \)) and structural (\( \eta \)) parameters, yielding the \textit{Causal Action Matrix} \( \mathcal{M} \), which encodes causal influence from states and past actions to future actions. In Phase 2 (\textit{Agent Planning with Causal Knowledge}), an LLM receives scenario \( t \) and proposes candidate actions \( \mathcal{A}' \). If \( \mathcal{A}' \neq \emptyset \), \textit{Causal-Aware Action Planning} adjusts LLM probabilities; if \( \mathcal{A}' = \emptyset \), \textit{Causal Backup Action} selects the most probable past action via \( \mathcal{M} \). Black solid arrows denote causal training; dashed arrows denote LLM inference, and red arrows denote causal knowledge consultation. The red box represents the causal score extraction for each potential next action, where the score is computed as the sum of causal contributions from the current state and previous action.
}
\label{fig:2_framework}  
\end{figure}

\subsection{Causal Action Structure Learning}
\subsubsection{Structural Causal Action (SCA) modeling}
\label{sect 3.3.1}

The goal of this step is to construct an SCA model, our extension of SCM, that captures the causal graph \( \mathcal{G} \), where the previous action \( a_{t-1} \) and the current state \( s_t \) are the parent nodes, and the next action \( a_t \) is the child node. Unlike prior work, which typically focuses on modeling state transitions or rewards~\citep{zhang2023interpretable,zhang2024causality}, our approach explicitly treats the action as a child node in the causal graph. This novel formulation allows the agent to reason causally about how past actions and current states influence future actions, providing a new perspective on decision-making dynamics. First we collect a dataset $B = \left\{ \{ (s_t^{k}, a_t^{k}) \}_{t=1}^T \right\}_{k=1}^N$
 using a behavior policy \(\pi_{\beta}\). We, then, factorize and discretely encode the states and actions into binary vectors:
\(
s_{t} = \left[ s_{t,1}, \dots, s_{t,S} \right] \in \{0,1\}^{S},
\quad
a_{t} = \left[ a_{t,1}, \dots, a_{t,A} \right] \in \{0,1\}^{A},
\)
where each component \(s_{t,j}\) and \(a_{t,i}\) is a binary indicator representing whether a particular state feature or action is active (1) or inactive (0) (refer to Appx.~\ref{sect A Causal Action Structure Learning Details} for details). The relationships between past states and actions that influence future actions can be represented as:

\begin{equation}
\label{eq 1}
a_i = f_i\left( \text{Pa}_{\mathcal{G}}(a_i), \varepsilon_{a_i} \right)
\end{equation}

for \(i=(1,2,\dots,A)\), where \( \text{Pa}_{\mathcal{G}}(a_i) \) denotes the parent nodes for \(a_i\) in the causal graph \(\mathcal{G}\). Inspired by previous research~\citep{ke2019learning,peng2022causality,zhang2023interpretable}, we define the function \( f_i \) as a neural network parameterized by the generating parameter \( \delta \). While the causal relationships of each graph are governed by the structural parameters encoded by binary adjacency indicators \(\eta_{ji}\). The loss function to optimize these parameters is: \(L(\delta,\eta)
= L_{\rm causal}(\delta,\eta)
+ L_{\rm reg}(\eta)
\). The causal loss function is:

\begin{equation}
\label{eq 3}
L_{\rm causal}
= \mathbb{E}_{(a_{t-1},s_t,a_{t})\sim B}
\Bigl[-\sum_{i=1}^{A}\log P\bigl(a_{t,i}\mid s_t,a_{t-1};\,\delta,\eta\bigr)\Bigr].
\end{equation}

A negative-log-prior penalty is imposed on the adjacency indicators to discourage spurious edges and avoid overfitting to unlikely causal links.  Let \(P(e_{ji}=1)\) be the prior probability for any edge.  Then
\begin{equation}
\label{eq 4}
L_{\rm reg}
= -\,\lambda\sum_{i,j}\eta_{ji}\,\log P\bigl(e_{ji}=1\bigr)
  \;,
\end{equation}

where \(\lambda>0\) controls the relative contribution of each penalty term.  Including an edge \(\eta_{ji}=1\) incurs a cost \(-\log P(e_{ji}=1)\), so only edges with high prior belief are preferred. In causal inference, identifiability, referring to the ability to recover causal effects from observed data~\citep{pearl2009causality} uniquely, is crucial for valid causal inferences. In our setting, identifiability guarantees that the true causal structure and decision policy can be recovered from observed trajectories. We can prove that under our formulation, identifiability is ensured, as stated in the following proposition:
\begin{proposition}[Identifiability]
Suppose that the state $s_t$ and previous action $a_{t-1}$ are observable, while the next action $a_t$ is unobservable, and they form a Markov Decision Process (MDP) as described in Eq.~\ref{eq 1}. Then, under the global Markov condition and the faithfulness assumption given a large enough dataset $B$, the next action $a_t$ is identifiable, as well as the causal structure characterized by the binary masks $\eta$ and the transition dynamics $f$.
\end{proposition}
\begin{proof}
See Appx.~\ref{Sect A Identifiability Analysis} 
\end{proof}

\textbf{Causal Action Matrix construction.}
We then construct the Causal Action Matrix \(
\mathcal{M} \) using the learned SCA model. The matrix $\mathcal{M}\in \mathbb{R}^{A \times (S+A)}$ encodes the causal score of selecting each action given the current state and past actions. Each row corresponds to a possible next action, and each column corresponds to a state or past action feature. Each entry \((i, j)\) of the matrix represents the probability that there is causal influence from state or action feature \(j\) to action \(i\), given by the learned structure parameter $\eta_{ji}$. Thus, a query $\mathcal{M}(s_t, a_{t-1}, a)$ will return the causal score $p_\text{c}(a)=\sum_{j \in J} \eta_{ji}$ where $J=\mathrm{Active}(s_t, a_{t-1}) \subseteq \{1, \dots, S+A\}$
denote the set of column indices corresponding to the features that are “active” in the current state \(s_t\) and the previous action \(a_{t-1}\) (details refer to Appx.~\ref{sect A Causal Knowledge Consultation}). To prevent cycles in the causal graph and ensure DAG property of a standard SCM~\citep{pearl2009causality}, we compare the coefficients for the bidirectional relationships in \(\mathcal{M} \) and set the lower one to \( 0 \).

\subsection{Agent Planning with Causal Knowledge}
During inference, we want to use the causal score from the matrix $\mathcal{M}$ to plan for the evaluation tasks. At time \( t \), the agent is given historical trajectory \( h_t = (s_1, a_1, s_2, a_2, \dots,a_{t-1},s_t)\). Instead of directly generating the next action \(a_{t+1}\), we want the agent first to consult the causal relationship given by $\mathcal{M}$. We sample from the LLM a set of actions, called \( \mathcal{A}' \subseteq \mathcal{A} \) for the scenario \( s_{t} \), which composes $\left\{ a'_{1}, a'_{2}, \dots, a'_{|\mathcal{A}'|} \right\}$
; this set $\mathcal{A}'$ contains actions that the agent believes to be correct. Each of these actions will come with a probability of being sampled by the backbone LLM, which we denote as \( p_{\text{a}}(a'_{m}) \). If \( \mathcal{A}' \neq \emptyset \) (there are actions in the instruction), we follow the process described in Sect.~\ref{sect 3.2.1} to modify the selection probability based on prior causal knowledge; otherwise, Sect. \ref{sect 3.2.2} to exploit the causal belief (none of the actions follow the instruction).

\subsubsection{Causal-Aware Action Planning}
\label{sect 3.2.1}
Given the sampled actions \( \mathcal{A}' \) with their associated probabilities \( P_{\text{a}}(\mathcal{A}') \), we aim to integrate the causal scores from the model \( \mathcal{M} \). We, first, extract the causal scores \(
p_{\text{c}}(a') = \mathcal{M}(s_t, a_{t-1}, a'), \forall a' \in \mathcal{A}',
\) (detail is described in Sect.~\ref{sect A Causal Knowledge Consultation}) to form the set
\(
P_{\text{c}}(\mathcal{A}') = \left\{ p_{\text{c}}(a'_{1}), p_{\text{c}}(a'_{2}), \dots, p_{\text{c}}(a'_{|\mathcal{A}'|}) \right\}
\).
The individual final action probabilities are computed as:

\begin{equation}
\label{eq:6}
p_{\text{f}}(a'_{m}) =  \gamma \cdot p_{\text{a}}(a'_{m}) + (1 - \gamma) \cdot p_{\text{c}}(a'_{m}),
\end{equation}

where $\gamma$ is a hyperparameter. We apply the softmax function to all value of $p_{\text{f}}(a'_{m})$ to  normalize the probabilities, which allows us to get the final probability set:

\begin{equation}
P_{\text{f}}(\mathcal{A}') = \left\{ p_{\text{f}}(a'_{1}), p_{\text{f}}(a'_{2}), \dots, p_{\text{f}}(a'_{|\mathcal{A}'|}) \right\}, \quad \sum_{k=1}^{|\mathcal{A}'|} p_{\text{f}}(a'_{m}) = 1
\end{equation}

The sampled action set \( \mathcal{A}' \) from the LLM agent may contain redundant actions, so we apply a method (detailed in Appx.~\ref{sect A Identifying Redundant Actions}) to identify and merge these duplicates by summing their probabilities. This yields a reduced set \( \mathcal{A}'^* \) with updated probabilities \( P^*_{\text{f}} \). This allows us to sample the next action:
    \begin{equation}
        a_{t} \sim \text{Categorical} \left( \left[ p^*_{\text{f}}(a'_{1}), p^*_{\text{f}}(a'_{2}), \dots, p^*_{\text{f}}(a'_{|\mathcal{A'^*}|}) \right] \right).
    \end{equation}

\subsubsection{Causal Backup Action }
\label{sect 3.2.2}
In the second case, where \( \mathcal{A}' = \emptyset \) — indicating that the agent samples actions \( a_t \notin \mathcal{A} \) — existing methods often apply soft interventions by prompting the agent to pause and replan~\citep{zhang2024proagent}. However, such strategies may fail when the agent persistently hallucinates invalid actions. Inspired by human behavior under uncertainty, we propose a recovery mechanism that leverages past causality knowledge. Instead of immediately replanning, the agent retrieves previously executed actions by querying \(
p_{\text{c}}(a) = \mathcal{M}(s_t, a_{t-1}, a), \forall a \in \mathcal{A},
\) (detail is described in Sect.~\ref{sect A Causal Knowledge Consultation}). This yields a probability distribution over the instructed action set \( \mathcal{A} \):
\(
P_{\text{c}}(\mathcal{A}) = \left\{ p_{\text{c}}(a_{1}), p_{\text{c}}(a_{2}), \dots, p_{\text{c}}(a_{A}) \right\}
\). Given the need to select the most reliable action during recovery based on the agent's past experiences thereby avoiding the potential risks of sampling an invalid action, we adopt a greedy strategy given by: 

\begin{equation}
a_{t} = \underset{a \in \mathcal{A}}{\arg\max} \ P_{\text{c}}(a).
\end{equation}

%% file: input/experiments.tex
\subsection{Experimental setup}
\label{sect 5.1}
We use the Overcooked-AI environment suite as our testing platform~\citep{carroll2019utility}. This suite comprises five distinct environments: \textit{Cramped Room} (CR), \textit{Asymmetric Advantages} (AA), \textit{Coordination Ring} (COR), \textit{Forced Coordination} (FC), and \textit{Counter Circuit} (CC), each presenting varying levels of difficulty. In these environments, two agents must collaborate to prepare and serve onion soup. Their tasks include gathering and placing up to three ingredients into a pot, cooking the soup, transferring it into a dish, and delivering the final meal (details of environments can be found in Appx.~\ref{sect A Environment Details}).  Both agents' actions contribute to a shared final return (successfully delivering the soup grants a reward of +20), making this environment well-suited for evaluating agent collaboration. Our experiments aim to demonstrate that the CausalPlan can enhance planning and, thus better collaboration for various open-sourced LLMs. Specifically,  we use \texttt{gemma-1.1-7b-it} (Gemma-7B), \texttt{Meta-Llama-3-8B-Instruct} (Llama-8B), \texttt{Qwen2.5-14B-Instruct-1M} (Qwen-14B), and \texttt{Llama-3.3-70B-Instruct} (Llama-70B) as the backbone LLMs to generate the action. We integrate these open-sourced LLMs into the ProAgent framework~\citep{zhang2024proagent} and apply CausalPlan to refine the planned actions. Additionally, we use \texttt{Cohere/command-r}~\cite{cohere2024commandr}, a 35-billion-parameter model, via the Cohere API to generate the analysis of the scenario for faster inference in our two-prompt input structure (refer to Sect.~\ref{sect A Detail Implementations} for details). Notably, the largest model that we used only has 70 billion parameters, rather small compared to larger GPT models that have been used in ProAgent~\citep{zhang2024proagent}.

In Sect. \ref{sect 4.2.1}, we compare the performance of LLM agents without CausalPlan to their performance when enhanced with CausalPlan. Our agent is evaluated alongside different AI agents (see next paragraph), following a common practice established in prior research~\citep{li2023cooperative, zhang2024proagent}. In these experiments, our agents play as Player 1 and the baseline agents as Player 0. An effective agent should demonstrate strong performance in collaboration with all other baseline agents.  
We also compare our CausalPlan agent with Llama-70B backbone against the baseline agents. 
In Sect.~\ref{sect 4.3}, we evaluate the performance of CausalPlan agents when collaborating with human-like agents (collected using Behavior Cloning proposed in~\citet{li2023cooperative}). In Sect.~\ref{sect 5.4}, we evaluate different components of CausalPlan, and in Appx. Sect.~\ref{sect A Additional Experiments}, we provide additional experiments such as parameter tuning, different data collection policies $\pi_\beta$, time complexity analysis, and the causal matrix found $\mathcal{M}$.

\textbf{Baselines.}
The baselines include traditional RL methods designed for zero-shot human and AI coordination. These baselines have achieved notable results in the field including SP~\citep{tesauro1994td, carroll2019utility}, PBT~\citep{jaderberg2017population}, FCP~\citep{strouse2021collaborating}, MEP~\citep{zhao2023maximum}, COLE~\citep{li2023cooperative} (refer to Appx.~\ref{sect A Baseline Details} for baselines' details).

\subsection{AI player evaluation}
\label{sect 4.2.1}
\begin{figure}[tbp]
\begin{center}
    \includegraphics[width=0.9\textwidth, height=3.5cm, trim={200pt 0pt 140pt 0pt}, clip]{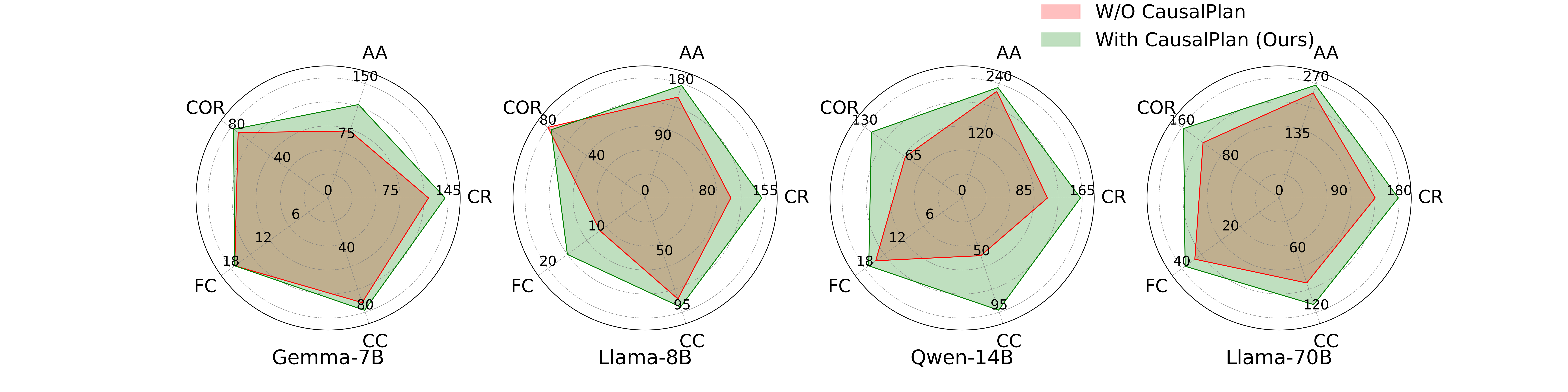}  
\end{center}
\caption{Performance of different backbones with and without CausalPlan across various layouts. In these experiments, we use the LLM agent as Player 1, allowing it to collaborate with all other baselines (described in Sect.~\ref{sect 5.1}) for 400 timesteps and report the average of three different seeds. Details are provided in Appx.~\ref{sect A Human-AI Evaluation} Tab.~\ref{tab:table A1}.}
\label{fig:backbone_compare}  
\end{figure}

\textbf{Enhancing open-sourced LLM performance using CausalPlan.}
We evaluate whether CausalPlan enhances open-sourced LLMs in collaboration tasks, as shown in Fig.~\ref{fig:backbone_compare} and detailed in Appx.~\ref{sect A Human-AI Evaluation}~Tab.~\ref{tab:table A1}.  CausalPlan improves performance across layouts and models, with significant gains seen in Qwen-14B (29.04\%) and Llama-70B (22.42\%). In terms of task layouts, the most substantial improvements were found in the CR (20.83\%) and COR (19.13\%) settings. Furthermore, CausalPlan also provided notable benefits for larger LLMs, such as Llama-70B, demonstrating its potential to enhance performance even at scale. For reference, the last row of Appx.~Tab.~\ref{tab:table A1} includes the results of using GPT-4 as the LLM agent, as reported in the ProAgent article~\citep{zhang2024proagent}. While we do not compare our method against this GPT-4 agent, due to the large size gap, we include this to illustrate how closely our method approaches oracle-level performance.

\textbf{Comparison with state-of-the-art RL baselines.}
We assess the performance of our top-performing agent (Llama-70B backbone) against the set of SOTA baseline RL agents. The results, presented in Tab.~\ref{tab:table 1}, show that our agent consistently ranks among the top performers across different layouts (highest score in three out of five layouts and ranks second in one additional layout). The most significant performance gaps between our method and the next best baseline are observed in the AA layout, showing a 63\% advantage. We attribute the underperformance in CR to the simplicity of the task, which does not demand causal knowledge. These results demonstrate that, when equipped with CausalPlan, open-source LLM agents can outperform state-of-the-art RL agents across various tasks, highlighting the effectiveness of integrating causal reasoning into LLM-based agents.

\begin{table}[t]
\caption{Average performance (mean ± std) of baseline agents and CausalPlan (Ours) across layouts using Llama-70B. Results are averaged over both player positions and three seeds (400 timesteps each). Best and second-best results are in \textbf{bold} and \underline{underlined}, respectively. Detailed performance of playing as Player 0 or Player 1 is provided in Appx.~Tab.~\ref{tab:table A1}.}
\label{tab:table 1}
\centering
\small
\renewcommand{\arraystretch}{1.2}
\begin{tabular}{l|c|c|c|c|c|c}
\hline
\multirow{2}{*}{\textbf{Layout}} & \multicolumn{5}{c|}{\textbf{Baseline AI Agents}} & \multirow{2}{*}{\textbf{\shortstack{CausalPlan \\ (Ours)}}} \\ \cline{2-6}
                                  & \textbf{SP} & \textbf{PBT} & \textbf{FCP} & \textbf{MEP} & \textbf{COLE} & \\ \hline
\textbf{CR}  & 162.0 $\pm$ 10.0 & \underline{168.0 $\pm$ 5.0} & \textbf{194.0 $\pm$ 10.1} & 178.0 $\pm$ 16.1 & 153.4 $\pm$ 12.5 & 172.7 $\pm$ 4.2 \\
\hline
\textbf{AA}  & 184.0 $\pm$ 17.5 & 168.0 $\pm$ 15.4 & 176.6 $\pm$ 15.0 & 167.3 $\pm$ 5.8 & \underline{185.3 $\pm$ 15.1} & \textbf{258.7 $\pm$ 16.4} \\
\hline
\textbf{CC}  & 56.7 $\pm$ 9.2  & 52.0 $\pm$ 14.0  & 63.4 $\pm$ 10.5  & 50.0 $\pm$ 16.1  & \underline{90.6 $\pm$ 10.1} & \textbf{112.6 $\pm$ 7.6} \\
\hline
\textbf{COR} & 120.7 $\pm$ 11.0 & 139.4 $\pm$ 10.1 & 130.7 $\pm$ 6.2 & \textbf{160.7 $\pm$ 7.2} & 153.4 $\pm$ 4.6 & \underline{156.6 $\pm$ 3.2} \\
\hline
\textbf{FC}  & 18.0 $\pm$ 4.6  & 40.6 $\pm$ 10.3  & \underline{42.0 $\pm$ 7.2}  & 30.4 $\pm$ 5.4  & 44.6 $\pm$ 7.0 & \textbf{53.9 $\pm$ 14.9} \\
\hline
\end{tabular}
\end{table}

\subsection{Human player evaluation}
\label{sect 4.3}
\begin{figure}[t]
\begin{center}
    \includegraphics[width=0.9\textwidth,height=4.3cm]{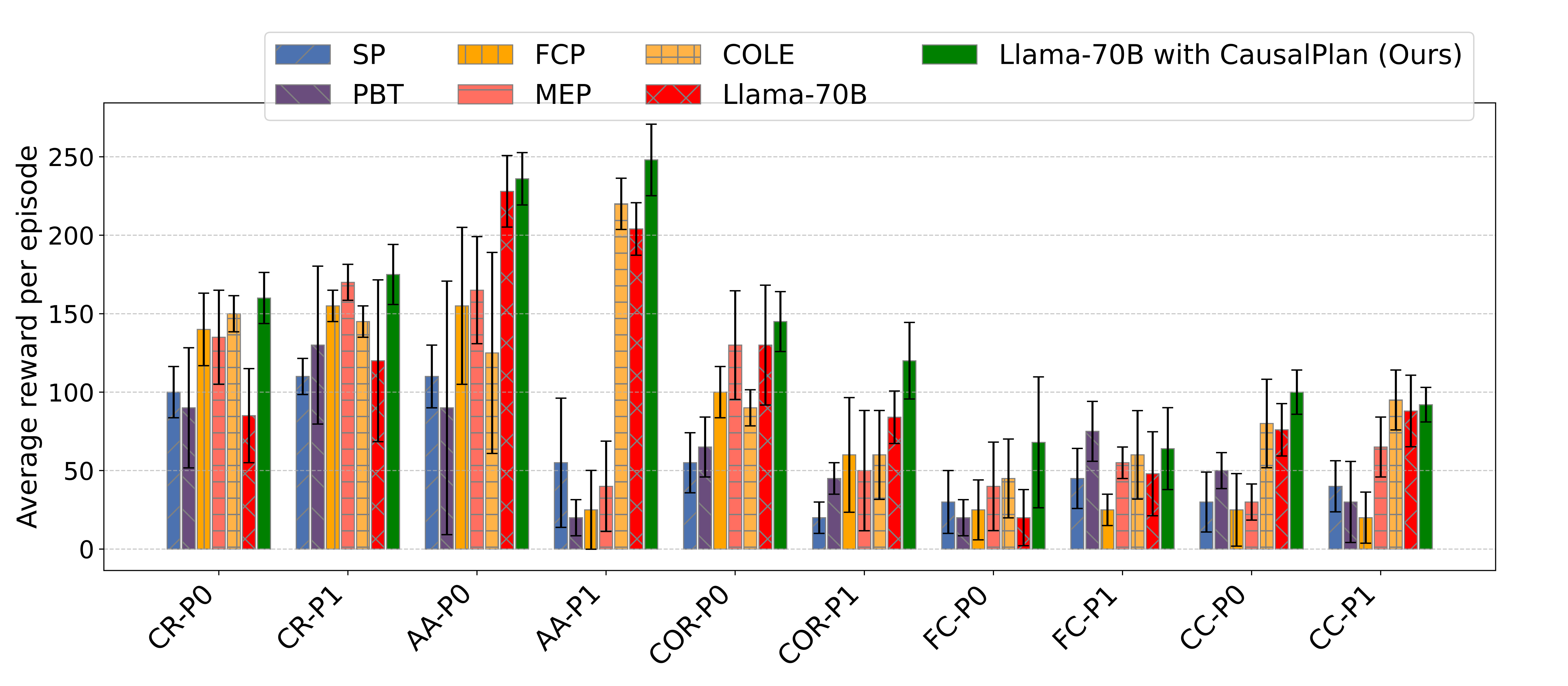}  
\end{center}
\caption{Experiments with a human proxy partner. Results show the mean and variance averaged over using five different BC policies as the partner (each running for 400 timesteps). "P0" denotes the controlled AI agent acting as Player 0, and vice versa.}
\label{fig:human_compare}  
\end{figure}

To evaluate human collaboration, we performed an experiment using human proxy partners, with the results shown in Fig.~\ref{fig:human_compare}. In this experiment, our CausalPlan framework utilizes Llama-70B as the backbone LLM. As shown, our agent (green bars) outperforms all baselines in 8 out of 10 configurations. On average across all layouts, it achieves approximately a 30\% improvement over the LLM agent without CausalPlan (red bars), and outperforms the strongest RL baseline (COLE) by approximately 32\%.

\subsection{Impact of CausalPlan components}
\label{sect 5.4}

\begin{table}[t]
\caption{Ablation studies were conducted on the CR layout using Llama-8B. "1-Prompt" uses a single prompt for observation and planning, as in ProAgent; "2-Prompt" uses our modified dual-prompt method (refer to Appx.~\ref{sect A "1P" vs "2P"}). "CausalPlan (no CBA)" omits the Causal Backup component.}
\label{tab:table 2}
\centering
\small
\renewcommand{\arraystretch}{1.2}  
\setlength{\tabcolsep}{2pt}  
\begin{tabular}{l|c|c|c|c|c|c}
\hline
\multirow{2}{*}{\textbf{Methods}} & \multicolumn{5}{c|}{\textbf{Baseline AI Agents}} & \multirow{2}{*}{\shortstack{\textbf{Average} \\ \textbf{Results}}} \\ \cline{2-6}

                                 & SP & PBT & FCP & MEP & COLE & \\ \hline
\textbf{1-Prompt (ProAgent)}                       & 86.7 $\pm$ 41.6 & 66.7 $\pm$ 63.4 & 180.0 $\pm$ 20.0 & 106.7 $\pm$ 75.7 & 113.3 $\pm$ 11.5 & 110.7 $\pm$ 12.8\\ \hline
\textbf{2-Prompt}            & 73.3 $\pm$ 30.5 & 93.3 $\pm$ 57.7 & 180.0 $\pm$ 0.0 & 126.7 $\pm$ 11.5 & 126.7 $\pm$ 23.1 &  121.3 $\pm$ 2.3\\ \hline
\textbf{CausalPlan (no CBA)}       & 113.3 $\pm$ 23.1 & 146.7 $\pm$ 46.2 & 160.0 $\pm$ 34.6 & 133.3 $\pm$ 11.5 & 153.3 $\pm$ 23.1 & 141.3 $\pm$ 12.9 \\ \hline
\textbf{CausalPlan (Full)}             & 126.7 $\pm$ 30.6 & 133.3 $\pm$ 30.5 & 160.0 $\pm$ 40.0 & 166.7 $\pm$ 41.6 & 166.7 $\pm$ 23.1 & \textbf{150.7 $\pm$ 2.3} \\ \hline
\end{tabular}
\end{table}
In this section, we investigate the individual contributions of each component within the CausalPlan framework and assess their impact on overall performance. First, we compare the use of a single prompt~\citep{zhang2024proagent}, for both observation analysis and planning against our two-prompt setup, where one prompt is dedicated to analysis and the other to planning (as described in Sect.~\ref{sect: Method}). This comparison helps isolate whether performance gains come from the embedded causal knowledge. As shown in Tab.~\ref{tab:table 2}, the performance between the single-prompt and two-prompt configurations is nearly identical, with only a slight improvement when using our two-prompt. This supports our claim that improvements primarily arise from causal reasoning. Second, we examine the effect of the Causal Backup Action module. CausalPlan without the backup action still outperforms the two-prompt variant by 27\%, but falls short of the full framework by 7\%. This highlights the significance of the backup mechanism to avoid scenarios in which the agent fails to select actions as instructed.

%% file: input/conclusion.tex
In this paper, we introduce \textbf{CausalPlan}, a framework designed to integrate causal reasoning into the planning processes of large language model (LLM) agents, with the goal of enhancing their performance in collaborative tasks. CausalPlan operates in two main phases. Our experiments show notable performance gains across various LLM backbones in collaboration settings. Despite these improvements, the framework still falls short of the performance achieved by larger models on certain tasks—a limitation we aim to address in future work. Additionally, due to limited access, we have not yet evaluated the method on larger closed-source LLMs, which represents an important direction for future research. This work serves as an initial step toward incorporating causal reasoning into multi-agent planning with language models. While the framework is not currently intended for deployment in specific applications, it holds the potential to improve the safety, efficiency, and interpretability of collaborative AI systems.

%% file: input/Appendix.tex
\section{Identifiability Analysis}
\label{Sect A Identifiability Analysis}
Proposition 1 (Identifiability of the causal structure and
 functions):

Let the dataset consist of sequences of the form: 
\begin{equation}
\label{eq: 8}
(s_t, a_{t-1}, a_t), \quad t = 1, \ldots, T,
\end{equation}

Assume the data comes from a Markov Decision Process (MDP) under the interaction of a fixed behavior policy $\pi_\beta$. The next action \(a_t\) (a binary vector of size \(A\)) is assumed to be generated by a structural causal model (SCM):

\begin{equation}
   a_{i,t} = f_i(\mathrm{Pa}(a_{i,t})) + \varepsilon_{a_{i,t}}, \quad i = 1, \ldots, A, 
\end{equation}

where, the parents $\mathrm{Pa}(a_{i,t})$ of $a_{i,t}$ are selected from the state $s_t$ and action $a_{t-1}$. The noise terms, $\varepsilon$, are independent of the parents. Under the following assumptions:

\begin{enumerate}
    \item \textbf{Additive noise}: The noise terms are independent and identically distributed (i.i.d.) and do not depend on the inputs~\cite{hoyer2008nonlinear}.
    \item \textbf{Causal sufficiency}: All relevant causes are observed (i.e., no hidden confounders)~\cite{spirtescausation}.
    \item \textbf{Faithfulness and global Markov condition}: Observed conditional independencies match those implied by the graph~\cite{pearl2009causality}.
    \item \textbf{Function class expressiveness}: Each function \(f_i\) belongs to a class identifiable under additive noise models. In additive noise models, identifiability of causal direction relies on the function class having sufficient expressiveness and satisfying certain regularity conditions (e.g., nonlinearity, invertibility)~\cite{ke2019learning,peters2017elements}.
    \item \textbf{Acyclicity}: The causal graph has no cycles (i.e., it is a Directed Acyclic Graph (DAG)).
    \item \textbf{Sufficient data}: There are enough samples to guarantee reliable estimation.
\end{enumerate}

Then, both the structure of the causal graph and the functions \(f_i\) can be identified. In particular, the binary adjacency masks indicating causal edges can be consistently estimated.

\textbf{Proof sketch:}

\textit{Step 1: Identifiability using additive noise models.} Under the above assumptions, especially additivity and faithfulness, each causal function \(f_i\) can be learned uniquely up to Markov equivalence. Prior work~\cite{hoyer2008nonlinear} shows that additive noise and independence of noise from inputs imply identifiability of the direction of causality.

\textit{Step 2:  Estimating the functions.} We approximate each function \(f_i\) using a weighted basis expansion:

\begin{equation}
f_i(\cdot) \approx W_i^\top \phi_i(\cdot),
\end{equation}

where \( \phi_i(\cdot) \in \mathbb{R}^d \) is a nonlinear feature map that transforms the input tuple \(( \cdot )\) into a \(d\)-dimensional representation, and \( W_i \in \mathbb{R}^d \) is the corresponding weight vector of function \( f_i \). In the simplest case, \(\phi_i(\cdot)\) and \(W_i\) are predefined basis functions and linear coefficients, respectively. However, in practice, we often implement \(f_i\) using a neural network to allow for flexible function approximation. Given an input tuple \( (s_t,a_{t-1}) \), the generating function for \( a_{i,t} \) can be rewritten as:

\begin{equation}
a_{i,t} = f_i(s_t, a_{t-1}) + \varepsilon_{a_{i,t}}
\approx W_i^\top \phi_i(s_t, a_{t-1}) + \varepsilon_{a_{i,t}},
\end{equation}

with noise term $\varepsilon_{a_{i,t}}$. 
 Suppose we have a dataset comprising \( N \) trajectories $k$ with the form given in Eq.~\ref{eq: 8}. For each trajectory, we define:

\begin{equation}
\Phi_i^k
=\begin{bmatrix}
\phi_i(s_1^{k},a_0^{k})^\top \\
\vdots\\
\phi_i(s_T^{k},a_{T-1}^{k})^\top
\end{bmatrix}
\in\mathbb{R}^{T\times d},
\qquad
\mathbf{A}^k_i
=\begin{bmatrix}
a_{i,1}^k\\
\vdots\\
a_{i,T}^k
\end{bmatrix}
\in\mathbb{R}^T,
\end{equation}

Each row of \( \Phi_i^k \) represents the feature vector for a specific time step, while the corresponding element in \( \mathbf{A}^k_i \) contains the observed action component. We then estimate \(W_i\) by minimizing the ridge‐regularized least‐squares objective:
\begin{equation}
\min_{W_i}
\sum_{k=1}^{N}
\bigl\|
\mathbf{A}^k_i \;-\;\Phi_i^k\,W_i
\bigr\|_2^2
\;+\;\lambda\,\bigl\|W_i\bigr\|_2^2,
\end{equation}
where \(\lambda > 0\) controls the regularization strength.

\textit{Step 3: Unique closed-form solution proof.} We proceed by proving that solving the objective yields a unique closed-form solution. The objective function above can be compactly written as:
\begin{equation}
L(W_i)
= \sum_{k=1}^N \bigl\|\mathbf{A}^k_i - \Phi_i^k W_i\bigr\|_2^2
\;+\;\lambda\,\|W_i\|_2^2
= \|\mathbf{A}_i - \Phi_i W_i\|_2^2 + \lambda\,W_i^\top W_i.
\end{equation}

First, expand the squared‐error term:
\begin{equation}
\|\mathbf{A}_i - \Phi_i W_i\|_2^2
= (\mathbf{A}_i - \Phi_i W_i)^\top(\mathbf{A}_i - \Phi_i W_i)
= \mathbf{A}_i^\top\mathbf{A}_i - 2\,W_i^\top\Phi_i^\top\mathbf{A}_i + W_i^\top\Phi_i^\top\Phi_i\,W_i.
\end{equation}

Thus,
\begin{equation}
L(W_i)
= \mathbf{A}_i^\top\mathbf{A}_i
-2\,W_i^\top\Phi_i^\top\mathbf{A}_i
+W_i^\top\Phi_i^\top\Phi_i\,W_i
+\lambda\,W_i^\top W_i.
\end{equation}

Taking the gradient with respect to \(W_i\) gives:
\begin{equation}
\nabla_{W_i} L(W_i)
= -2\,\Phi_i^\top\mathbf{A}_i + 2\,\bigl(\Phi_i^\top\Phi_i + \lambda I\bigr)\,W_i.
\end{equation}

Setting \(\nabla_{W_i} L(W_i) = 0\) yields the normal equation:
\begin{equation}
\bigl(\Phi_i^\top\Phi_i + \lambda I\bigr)\,W_i
= \Phi_i^\top\mathbf{A}_i.
\end{equation}

Since \(\lambda > 0\), the matrix \(\Phi_i^\top\Phi_i + \lambda I\) is strictly positive‐definite and hence invertible. Therefore, the unique minimizer is:
\begin{equation}
\boxed{
W_i
= \bigl(\Phi_i^\top\Phi_i + \lambda I\bigr)^{-1}\,\Phi_i^\top\mathbf{A}_i.
}
\end{equation}

Re‐expressing in terms of the individual trajectories,
\begin{equation}
\Phi_i^\top\Phi_i = \sum_{k=1}^N (\Phi_i^k)^\top\Phi_i^k,
\quad
\Phi_i^\top\mathbf{A}_i = \sum_{k=1}^N (\Phi_i^k)^\top \mathbf{A}^k_i,
\end{equation}
so equivalently
\begin{equation}
W_i
= \Bigl(\sum_{k=1}^N (\Phi_i^k)^\top\Phi_i^k + \lambda I\Bigr)^{-1}
\sum_{k=1}^N (\Phi_i^k)^\top \mathbf{A}^k_i.
\end{equation}

Because \(L(W_i)\) is strictly convex, it admits a unique closed-form solution. Moreover, given a sufficiently large dataset, the estimator converges to a good estimate of \(W_i\)~\cite{williams2006gaussian}.

\subsection*{Step 3: Recovering the graph structure}

To recover the graph structure, we exploit the closed-form solution for \(W_i\) derived by minimizing the regularized quadratic loss in the previous step:
\[
W_i = \left( \Phi_i^\top \Phi_i + \lambda I \right)^{-1} \Phi_i^\top \mathbf{A}_i.
\]
This expression yields an estimate of the weight vector \(W_i\), which quantifies the linear relationship between the current state and previous action features and the target component \(a_{i,t}\). The \emph{support} of \(W_i\)—i.e., the indices of its nonzero entries—identifies which features are informative for predicting \(a_{i,t}\). Under the \emph{faithfulness} assumption, this support exactly corresponds to the true parent set of node \(i\) in the underlying causal graph. Thus, one can recover the graph structure by examining which entries of \(W_i\) are significantly nonzero, using thresholding or statistical tests.

\subsection*{Conclusion}

Under the usual identifiability conditions, both the graph structure and the functional relationships in the Structural Causal Action model are uniquely determined.  As a result, the causal action matrix learned by CausalPlan faithfully reflects the true cause–effect relations among states and actions.

\section{CausalPlan Details}

\label{sect A CausalPlan Details}
As described in Sect.~\ref{sect: Method} and in Fig.~\ref{fig:2_framework}, the CausalPlan framework involves two phases Causal Action Structure Learning and Agent Planning with Causal Knowledge. Here, we discuss in detail the two phases and their components, as well as present algorithms that outline the method.

\subsection{Causal Action Structure Learning details}
\label{sect A Causal Action Structure Learning Details}
This appendix outlines the procedure used to model and learn the causal relationships between the previous action \( a_{t-1} \), current state \( s_t \), and next action \( a_t \).

\textbf{Buffer \( B \) Collection.} The data collection process begins by constructing the buffer \( B \), which is used to train the SCA model. We collect this data by allowing a pretrained agent to interact with the environment for \( N \) timesteps, with each episode having a horizon of \( T \). These interactions include both high-level task-oriented actions and low-level movement actions. 

To facilitate causal analysis, we apply a preprocessing step in which all low-level movement actions are relabeled as the most recent preceding high-level action of interest. For example, if the agent executes \texttt{pickup\_onion}, then moves for several steps, and finally performs \texttt{put\_onion\_in\_pot}, all intermediate movement actions are relabeled as \texttt{pickup\_onion}. This yields a simplified sequence: \texttt{pickup\_onion} \( \rightarrow \) \texttt{pickup\_onion} \( \rightarrow \) \texttt{pickup\_onion} \( \rightarrow \) \texttt{put\_onion\_in\_pot}. This transformation reduces noise from irrelevant actions and makes it easier to detect meaningful causal edges—such as from \texttt{pickup\_onion} to \texttt{put\_onion\_in\_pot}.

Importantly, we retain the original state observations at each timestep, even after relabeling the actions. This ensures that we can still study the causal relationship between the immediate state before an action and the subsequent high-level decision, preserving the integrity of the underlying state-action dynamics.

\textbf{SCA Model.} To capture these dependencies, we employ the SCA model, which incorporates two key components: the generative parameters \( \delta \) and the structural parameters \( \eta \). The parameters \( \delta \) define a set of functions \( f \), each implemented as a neural network. Specifically, for each action feature \( a_i \) in Eq.~\ref{eq 1}, there is a corresponding function \( f_i \) parameterized by \( \delta_i \) (see Appx.~\ref{sect A Hyperparameters} for network details). As described in Sect.~\ref{sect 3.3.1}, the model generates the next action \( a_t \) based on the current state and previous action. The parameters \( \delta \) govern this generative mapping and are trained using standard neural network optimization.

In parallel, the structural parameters \( \eta \) encode the causal graph \( \mathcal{G} \), where each entry indicates the presence or absence of a directed edge between action factorizations, using binary adjacency indicators. During inference, only those factorizations that are parents of the current action feature \( a_i \), according to \( \eta \), are activated. We implement this by masking out all features not connected to \( a_i \), ensuring that each function \( f_i \) conditions only on its relevant causal parents, as defined in Eq.~\ref{eq 1}. Furthermore, we manually set the diagonal entries \( \eta_{i \to i} = 0 \), since edges from an action factorization to itself are not allowed in the causal graph. This constraint prevents self-causation among action nodes, maintaining a valid causal structure. Finally, we apply a sigmoid activation to each entry of \( \eta \), producing values in \([0, 1]\) that represent the probability of an edge's existence

\textbf{Optimization.} Both sets of parameters are jointly optimized during training. The overall loss function is defined as
\[
L(\delta, \eta) = L_{\text{causal}}(\delta, \eta) + L_{\text{reg}}(\eta),
\]
where \( L_{\text{causal}} \) encourages accurate prediction of the next action, and \( L_{\text{reg}} \) regularizes the structural parameters to promote sparsity and prevent overfitting. This results in an interpretable and reliable causal model.

The full training procedure is summarized in Algorithm~\ref{alg:1}, which alternates between updating the generative and structural parameters using mini-batches sampled from the buffer \( B \).

\begin{algorithm}
\caption{Iterative Optimization for Structural Causal Action (SCA) Model}
\label{alg:1}
\begin{algorithmic}[1]
\STATE \textbf{Input:} Dataset $B = \left\{ \{ (s_t^{k}, a_t^{k}) \}_{t=1}^T \right\}_{k=1}^N$
\STATE Initialize structural parameters \( \eta \) and generating parameters \( \delta \)
\STATE \textbf{Repeat:}
\STATE \quad \textbf{1. Sample a mini-batch} \( \mathcal{B} = \left\{ \{  (a^k_{t-1}, s^k_t, a^k_t)_{t \in T} \right\}_{k \in N} \subset B \)
\STATE \quad \textbf{2. Optimize Generating Parameters \( \delta \):}
\STATE \quad \quad \textbf{Fix} \( \eta \)
\STATE \quad \quad Optimize \( \delta \) by minimizing the loss \( L_{\rm causal}(\delta, \eta) \) in Eq.~\ref{eq 3}
\STATE \quad \quad Update generating parameters \( \delta \)
\STATE \quad \textbf{3. Optimize Structural Parameters \( \eta \):}
\STATE \quad \quad \textbf{Fix} \( \delta \)
\STATE \quad \quad Optimize \( \eta \) by minimizing the loss \(L(\delta,\eta)
= L_{\rm causal}(\delta,\eta) 
+ L_{\rm reg}(\eta)
\) in Eq.~\ref{eq 3} and Eq.~\ref{eq 4}
\STATE \quad Apply sigmoid to \( \eta \)
\STATE \textbf{Output:} Optimized parameters \( \delta, \eta \)
\end{algorithmic}
\end{algorithm}

\subsubsection{State and action factorization}
We assumed a known factorization of state and action spaces, a common assumption often made in causal reinforcement learning research~\citep{seitzer2021causal,peng2022causality}. This allows us to encode the states and actions into binary vectors: \(
s_{t} = \left[ s_{t,1}, \dots, s_{t,S} \right] \in \{0,1\}^{S},
\quad
a_{t} = \left[ a_{t,1}, \dots, a_{t,A} \right] \in \{0,1\}^{A},
\)
where each component \(s_{t,j}\) and \(a_{t,i}\) is a binary indicator representing whether a particular state feature or action is active (1) or inactive (0).

For example, given an observation $s^k_t$ of trajectory $k$ at timestep $t$ : ``agent 1 is holding an onion, agent 2 is holding nothing'', this can be encoded into a binary state vector such as:
\[
s^k_t = [1, 0, 0, 1, 0, \dots],
\]
where each entry corresponds to a specific feature (e.g., ``agent 1 is holding onion'', ``agent 1 is holding nothing'', ``agent 2 is holding nothing'', etc.), and the 1s indicate which conditions are currently true. Similarly, an action like ``agent 1 places onion in pot'' can be encoded into
\[
a^k_t = [0, 1, 0, \dots],
\]
where each entry corresponds to a specific atomic action in the action space, and the 1 marks the active action at time \( t \). \textbf{Note:} In our training process, we use only the previous action of the controlling agent. While it is possible to incorporate the actions of the other agent, doing so increases the complexity of learning the causal graph and may negatively impact the training performance.

This factorized representation enables us to formulate the causality training as a classification problem, allowing us to optimize using the negative log-likelihood loss defined in Eq.~\ref{eq 3}. Refer to Appx.~\ref{sect A States and actions factorization in each environment} for the factorization features used in our experiments.

\subsection{Agent Planning with Causal Knowledge details}
\label{sect A Agent Planning with Causal Knowledge Details}
This appendix provides additional details on how causal knowledge is integrated into the agent's decision-making process during action planning.

\textbf{LLM prompting process.} During inference, we first equip the LLM agents with a knowledge library that specifies the tasks, rules, and example responses relevant to the game environment. At each time step, the current observation \( s_t \) is presented to the agent along with a prompt instructing it to analyze the situation. The agent typically responds with a natural language interpretation highlighting the key elements of the observation. Both the original observation \( s_t \) and the generated analysis are then fed into a second prompt, which instructs the agent to produce a set of appropriate next actions \( \mathcal{A}' \). For further details, refer to Appx.~\ref{sect A "1P" vs "2P"}.

\textbf{Causal-Aware Action Planning.} When a set of candidate actions \( \mathcal{A}' \) is generated during planning, each action is initially assigned a probability by the LLM model, denoted as \( P_{\text{a}}(\mathcal{A}') \). To incorporate causal reasoning, the agent queries the Causal Action Matrix \( \mathcal{M} \) using the current state \( s_t \) and previous action \( a_{t-1} \) to compute a corresponding set of causal scores \( P_{\text{c}}(\mathcal{A}') \) (refer to Appx.~\ref{sect A Causal Knowledge Consultation} for details). A weighted combination of the LLM's probabilities and causal scores is formed using Eq.~\ref{eq:6} and then normalized via the softmax function:

\begin{equation}
\label{A eq 7}
  \frac{\exp(p_{\text{f}}(a'_{m}))}{\sum_{j=1}^{|\mathcal{A}'|} \exp(p_{\text{f}}(a'_{m}))},  
\end{equation}

resulting in the final action distribution \( P_{\text{f}}(\mathcal{A}') \). Redundant actions are identified and merged according to the process described in Appx.~\ref{sect A Identifying Redundant Actions}, and the agent samples the next action \( a_{t+1} \) from this refined distribution.

\textbf{Causal Backup Action.} In scenarios where no valid candidate actions are proposed (i.e., \( \mathcal{A}' = \emptyset \)), mostly due to hallucinations, the agent relies on a causal fallback mechanism. Instead of halting execution, it queries \( \mathcal{M} \) using \( s_t \) and \( a_{t-1} \) to derive a causal distribution over the original instruction set \( \mathcal{A} \). The agent then selects the action with the highest causal score, effectively leveraging prior experience to recover from failure.

The complete inference procedure using Causal-Aware Action Planning and Causal Backup Action is summarized in Algorithm~\ref{alg:2}.

\begin{algorithm}
\caption{Agent Planning with the Causal Knowledge Algorithm at time step \( t \)}
\label{alg:2}
\begin{algorithmic}[1]
\STATE \textbf{Input:} Current state \( s_t \), previous action \( a_{t-1} \), candidate actions \( \mathcal{A}' \), LLM probabilities \( P_{\text{a}}(\mathcal{A}') \), instruction set \( \mathcal{A} \), causal matrix \( \mathcal{M} \), weighting coefficient \( \gamma \in [0,1] \), \( P_\text{f}(\mathcal{A'}) = \emptyset \), \( P_c(\mathcal{A}) = \emptyset \) 
\STATE \textbf{If} \( \mathcal{A}' \neq \emptyset \) \textbf{then}
    \STATE \quad \textbf{For all} \( a'_m \in \mathcal{A}' \)
    \STATE \quad \quad $p_{\mathrm{c}}(a'_m) \leftarrow \mathcal{M}(s_t,a_{t-1},a'_m)$
    \STATE \quad \quad \( p_{\text{f}}(a'_m) \leftarrow \gamma \cdot p_{\text{a}}(a'_m + (1 - \gamma) \cdot p_{\text{c}}(a'_m) \) (Eq.~\ref{eq:6})
    \STATE \quad \quad \( P_\text{f}(\mathcal{A'})\leftarrow p_{\text{f}}(a'_m) \)
    \STATE \quad \textbf{End for} 
    \STATE \quad Normalize \( P_\text{f}(\mathcal{A'}) \) using softmax in Eq~\ref{A eq 7}
    \STATE \quad Apply redundancy check (see Appx.~\ref{sect A Identifying Redundant Actions}) to get \( \mathcal{A}'^* \), \( P_{\text{f}}^* \)
    \STATE \quad Sample $a_{t} \sim \text{Categorical} \left( \left[ p^*_{\text{f}}(a'_{1}), p^*_{\text{f}}(a'_{2}), \dots, p^*_{\text{f}}(a'_{|\mathcal{A'^*}|}) \right] \right)$
\STATE \textbf{Else}
    \STATE \quad \textbf{For all} \( a \in \mathcal{A} \)
        \STATE \quad \quad \( p_{\text{c}}(a) \leftarrow \mathcal{M}(s_t,a_{t-1},a) \)
        \STATE \quad \quad \( P_\text{c}(\mathcal{A})\leftarrow p_{\text{c}}(a) \)
    \STATE \quad \textbf{End for} 
    \STATE \quad \( a_{t} \leftarrow \arg\max_{a \in \mathcal{A}} \ P_{\text{c}}(a) \)
\STATE \textbf{End If}
\STATE \textbf{Output:} Selected action \( a_{t} \)
\end{algorithmic}
\end{algorithm}

\subsubsection{LLM prompt design}
\label{sect A "1P" vs "2P"}

\begin{figure}[H]
\begin{center}
    \includegraphics[width=1\textwidth, trim={0pt 30pt 0pt 500pt}, clip]{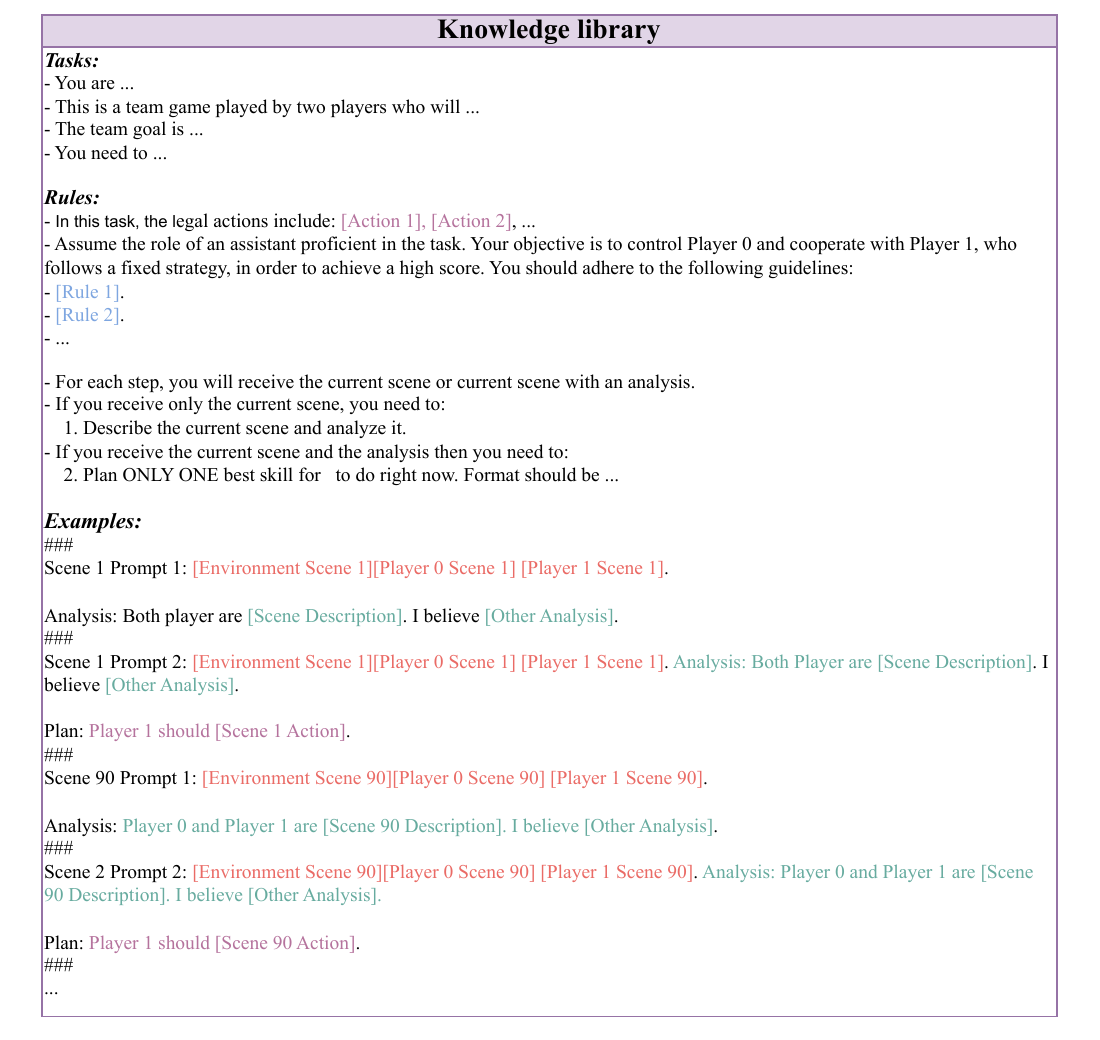}  
\end{center}
\caption{An Example of Knowledge Library.}
\label{fig:knowledge_library}  
\end{figure}

\textbf{Knowledge library.} At the beginning of the inference process, we construct a knowledge library for the LLM agent, following prior work in the field~\citep{zhang2024proagent,qiao2024agent}. This library is organized around three key perspectives: the tasks, the rules, and the in-context examples. This knowledge library is fed into the LLMs at the initial stage of the inference process before the cooperation task begins. An example of a knowledge library is provided in Fig.~\ref{fig:knowledge_library}.

In our experiments, for simplicity, we utilized the knowledge library provided by~\citet{zhang2024proagent}, with slight modifications to accommodate our two-prompt design, as their work uses the same evaluation environment\footnote{\url{https://github.com/PKU-Alignment/ProAgent} (MIT License).}.

\begin{figure}[H]
\begin{center}
    \includegraphics[width=1\textwidth, trim={0pt 30pt 0pt 250pt}, clip]{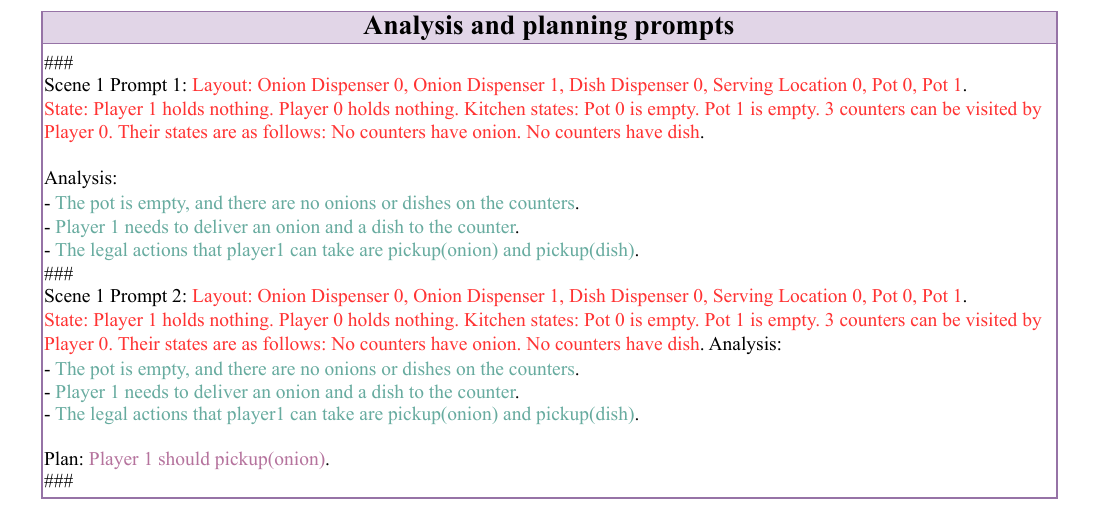}  
\end{center}
\caption{An example of analysis and planning prompts.}
\label{fig:analysis_and_planning_prompts}  
\end{figure}

\textbf{Analysis and planning prompts}
To facilitate the planning process, we first ground the environment state into natural language so that it becomes interpretable to the LLM agent, as the raw state representation is typically not directly understandable by language models. In our experiments, we adopt the grounding methodology proposed by~\citet{zhang2024proagent}, since their work uses the same evaluation environments. For detailed grounding procedures, we refer the reader to their paper. An example of the final grounded state prompt used as input to the agent at each timestep is highlighted in \textcolor{red}{red} in Fig.~\ref{fig:analysis_and_planning_prompts}.

We then apply our two-prompt design to guide the LLM’s behavior using the knowledge library. Specifically, when the agent is prompted with only the current observation, it is expected to analyze the scene. When the prompt includes both the observation and the analysis, the agent is expected to respond with a planned action. Our approach first asks the agent to perform the analysis, then uses that analysis together with the state prompt as input to generate the final action plan. The analysis is highlighted in \textcolor{customgreen}{green}, while the planned action is highlighted in \textcolor{custompurple}{purple} in Fig.~\ref{fig:analysis_and_planning_prompts}. We hypothesize that this two-prompt process provides the agent with a reasoning workflow similar to the chain-of-thought (CoT) prompting described by~\citet{wang2022self}, while also allowing straightforward access to the planned action through hard-coded separation. In contrast, including both the analysis and the planned action in the same response, as done by~\citet{zhang2024proagent}—can make it difficult to accurately extract the planned action, since action names might appear within the analysis. We evaluate the performance of one-prompt versus two-prompt approaches without causality enhancement through our CausalPlan in Sect.~\ref{sect 5.4} and find that the results are quite similar, with the two-prompt approach showing slightly better performance. Although the single-prompt approach is feasible in practice, it complicates reliably identifying the correct action.

\subsubsection{Causal knowledge consultation details}
\label{sect A Causal Knowledge Consultation}
To compute the causal score for a candidate action, the agent first maps the action to its corresponding row in \( \mathcal{M} \) and identifies which columns are currently active based on features derived from the current state and previous action. These active features are determined using the procedure outlined in Appx.~\ref{sect A Extracting Scenario Information for Causal Knowledge Consultation}. 

For instance, given that we want to extract the causal scores of an action \( a \), given current state \(s_t\) and previous action \(a_{t-1}\), we first identify the corresponding index \( i \) of the action \( a \)  within the matrix row.
Let
\(
\mathbf{idx}\colon \mathcal{A} \to \{1, \dots, |\mathcal{A}|\}
\)
be the function that maps any action to its row index in \(\mathcal{M}\), and let
\(
J=\mathrm{Active}(s_t, a_{t-1}) \subseteq \{1, \dots, S+A\}
\)
denote the set of column indices corresponding to the features that are “active” in the current state \(s_t\) and the previous action \(a_{t-1}\).

For a candidate action \(a\), we first compute its row index $i = \mathbf{idx}(a),$ then gather the entries of row \(i\) in \(\mathcal{M}\) at all active columns \(j \in J \), thus a query $\mathcal{M}(s_t,a_{t-1},a)$ will return:
\begin{equation}
p_{\mathrm{c}}(a) = \sum_{j \in J} \eta_{ji}.
\end{equation}


In other words, \(p_{\mathrm{c}}(a)\) is the sum of the causal-weight entries in the row for \(a_t\) that correspond to the features currently active.

\subsubsection{Extracting information for causal knowledge consultation}
\label{sect A Extracting Scenario Information for Causal Knowledge Consultation}
\begin{figure}[H]
\begin{center}
    \includegraphics[width=1\textwidth]{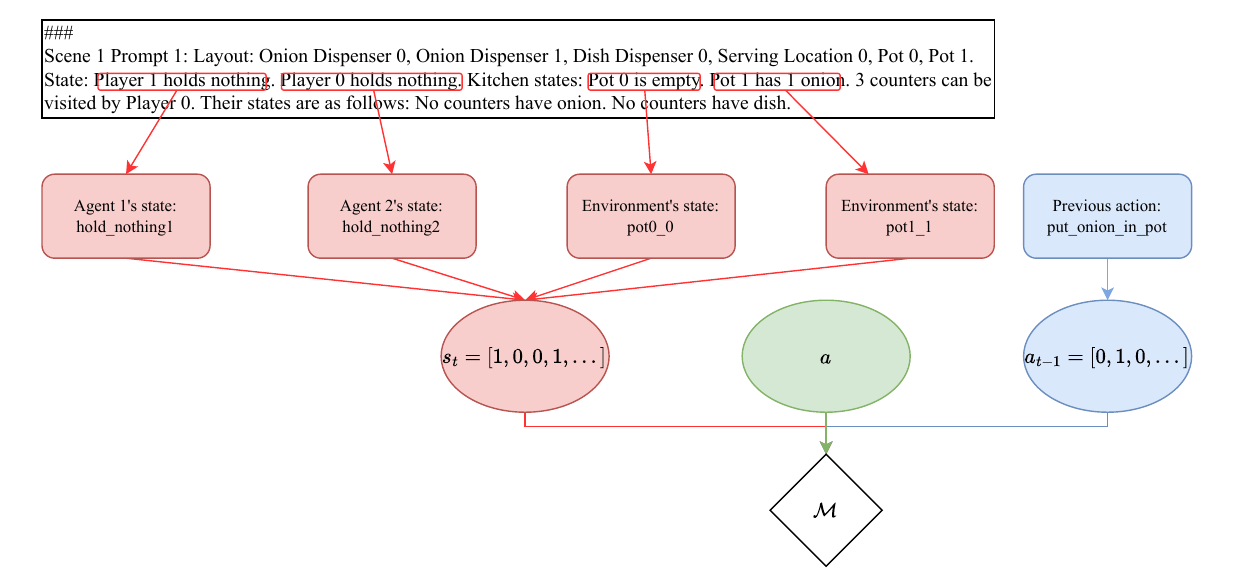}  
\end{center}
\caption{Information extraction for causal knowledge consultation.}
\label{fig:state_information_extraction}  
\end{figure}

Given the observations grounded in natural language, as explained in Appx.~\ref{sect A "1P" vs "2P"}, we map them to a set of predefined state features. For example, from the state prompt shown in Fig.~\ref{fig:state_information_extraction} --- ``Player 1 holds nothing. Player 0 holds nothing. Kitchen states: Pot 0 is empty. Pot 1 has 1 onion\ldots'' --- we extract factorized features such as \texttt{hold\_nothing1} (indicating that agent 1 is holding nothing), \texttt{hold\_nothing2} (agent 2 is holding nothing), \texttt{pot0\_0} (pot 0 is empty), \texttt{pot1\_1} (pot 1 has 1 onion). This allows us to formulate the state feature factorization $s_t$.

In addition, the previous action taken by the agent is also recorded, referring to the last executable action performed. This allows us to configure the action feature factorization vector $a_{t-1}$.

Depending on the environment, the number of factorized features can vary widely (see Appx.~\ref{sect A States and actions factorization in each environment} for the specific factorized features used in each experimental task). While a larger number of features can produce a more detailed causal graph, this does not necessarily lead to better performance, as learning such graphs becomes more challenging and requires more data. In our experiments, we chose to use high-level state and action factorizations (we ignore low-level movement actions and only focus on the state of the two agents and the pot) to strike a balance between expressiveness and learnability.

\subsubsection{Post-processing to identify redundant actions}
\label{sect A Identifying Redundant Actions}

During the process of sampling the next actions, the LLM may output the same action in different formats within the sampled set $\mathcal{A'}$. To address this, we apply a series of post-processing steps using standard natural language processing techniques—such as converting text to lowercase, removing punctuation, and regex matching pre-defined patterns—to identify and merge semantically equivalent actions. This enables us to accurately aggregate their probabilities in $P_\text{f}(\mathcal{A'})$. For instance, the same action \texttt{put\_onion\_in\_pot} can be expressed as \texttt{put\_onion\_in\_pot()}, \texttt{put\_onion\_in\_pot().}, or \texttt{put\_onion\_In\_Pot} (refer to the associated code for details of this process). After post-processing all these possible responses, we can calculate the updated value:

\[
\begin{aligned}
p_\text{f}(\texttt{put\_onion\_in\_pot}) &= p_\text{f}(\texttt{put\_onion\_in\_pot()}) \\
&\quad +\ p_\text{f}(\texttt{put\_onion\_in\_pot().}) + p_\text{f}(\texttt{put\_onion\_In\_Pot})
\end{aligned}
\]

\section{Additional Experiment Details}
\subsection{Additional implementation details}
\label{sect A Detail Implementations}
\subsubsection{CausalPlan implementation}
\label{sect A CausalPlan Implementation}

As mentioned earlier, we build upon the ProAgent framework~\citep{zhang2024proagent}, retaining all components except for the planning module, which we replace with our proposed algorithm. Unlike the original ProAgent implementation that relied on the closed-source \texttt{GPT-3.5}~\footnote{We would like to clarify a minor detail in the main manuscript that will be corrected in the revised version. The ProAgent framework was referred to as using \texttt{GPT-4}, based on the default model loaded in the code. However, according to the \texttt{README}, the results reported in the paper were actually obtained using \texttt{gpt-3.5-turbo-0301}. We will update the manuscript accordingly to reflect this.} for planning, we instead utilize one of the following open-source language models, all retrieved from Hugging Face\footnote{\url{https://huggingface.co}}: \texttt{gemma-1.1-7b-it} (Gemma-7B), \texttt{Meta-Llama-3-8B-Instruct} (Llama-8B), \texttt{Qwen2.5-14B-Instruct-1M} (Qwen-14B), and \texttt{Llama-3.3-70B-Instruct} (Llama-70B). These models are integrated into the ProAgent framework to serve as the core planner, with our \texttt{CausalPlan} method applied to refine the generated actions.  Additionally, for the two-prompt input structure, we employ the \texttt{Cohere/command-r} model~\citep{cohere2024commandr}---a 35-billion-parameter LLM accessed via the Cohere API using the official \texttt{cohere} Python client\footnote{\url{https://docs.cohere.com/v2/reference/chat}}---to produce scenario analyses for faster inference. For the ``Belief Correction'' module, we also substitute \texttt{GPT-3.5} with the same Cohere model. The "Controller" module in ProAgent~\citep{zhang2024proagent}—and in our setup—uses a rule-based best-first search; while effective, performance could likely be improved with a reinforcement learning-based approach.

Regarding hardware requirements, the Gemma-7B and Llama-8B models each require approximately 10--16~GB of VRAM, Qwen-14B demands around 25--30~GB and multi-GPU support, while Llama-70B needs over 70~GB VRAM with multi-GPU configuration on NVIDIA h-100 GPUs.

To facilitate easier extraction of action selection probabilities, we slightly modify the prompting strategy used in the original method. In particular, we separate the reasoning step, based on CoT prompting, from the action planning step, implementing them as two distinct prompts. The output of the reasoning prompt is then used as input for the planning prompt. We provide further details of this process in Appx.~\ref{sect A "1P" vs "2P"} and include an empirical study in Appx.~\ref{sect A Effect of Different Data Collection Agents}, demonstrating that this modification does not contribute to the performance gains, nor does it substantially affect the overall performance of the backbone. 

To avoid the cold-start problem and long interaction times associated with using small LLMs to collect data into the buffer $B$, we employ a pre-trained policy based on MEP to interact with the environment and gather data. Nonetheless, we conduct an experiment (results are in Appx.~\ref{sect A Effect of Different Data Collection Agents}) demonstrating that even when using a small LLM, specifically Llama-8B, for data collection, our method still yields improved performance compared to simply using the backbone method.

To access the open-source models

\subsection{Environment details}
\label{sect A Environment Details}
We use the Overcooked-AI environment suite as our testing platform~\citep{carroll2019utility}. In Overcooked, two agents must collaborate to prepare and serve onion soup. Their tasks include gathering and placing up to three ingredients into a pot, cooking the soup, transferring it into a dish, and delivering the final meal. Each successful delivery yields a reward of +20, and both agents share the final return, promoting cooperative behavior. This suite comprises five distinct layouts~\citep{carroll2019utility}—\textit{Cramped Room} (CR), \textit{Asymmetric Advantages} (AA), \textit{Coordination Ring} (COR), \textit{Forced Coordination} (FC), and \textit{Counter Circuit} (CC)—each designed to evaluate different aspects of multi-agent collaboration under varying levels of complexity and coordination demands: 

\begin{itemize}
    \item \textbf{Cramped Room (CR):} This environment features a highly constrained layout with narrow hallways and tight corridors, forcing agents to navigate around each other constantly.
    \item \textbf{Asymmetric Advantages (AA):} In AA, the kitchen layout provides one agent with easier access to ingredients and tools, while the other agent is disadvantaged in terms of spatial reach.
    \item \textbf{Coordination Ring (COR):} COR introduces a ring-like structure in the kitchen, where ingredients, cooking stations, and delivery points are spread along a loop.
    \item \textbf{Forced Coordination (FC):} FC is designed to enforce interdependence between the agents through environment constraints.
    \item \textbf{Counter Circuit (CC):} The CC environment includes a set of counters that create a barrier between the agents and the task stations.
\end{itemize}

\begin{figure}[H]
\begin{center}
    \includegraphics[width=1\textwidth]{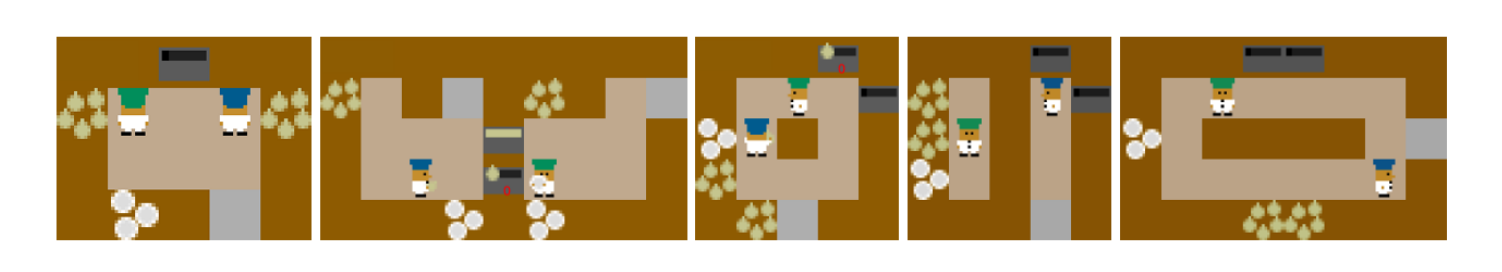}  
\end{center}
\caption{Overcooked-AI Environments. From left to right: Cramped Room (CR), Asymmetric Advantages (AA), Coordination Ring (CR), Forced Coordination (FC), and Counter Circuit (CC).}
\label{fig:Overcooked}  
\end{figure}

The environment testing suite was collected from the associated GitHub repository\footnote{\url{https://github.com/HumanCompatibleAI/overcooked_ai} (MIT License)}.

\subsection{Baseline details}
\label{sect A Baseline Details}
We compare CausalPlan against several established reinforcement learning (RL) methods specifically designed for zero-shot human-AI coordination tasks. These baselines have demonstrated strong performance in prior research and serve as competitive benchmarks in our experiments.

\begin{itemize}
    \item \textbf{SP (Self-Play)}~\citep{tesauro1994td, carroll2019utility}: A classical RL approach where agents learn policies by playing against themselves, promoting strategic behavior without relying on external partners.
    \item \textbf{PBT (Population-Based Training)}~\citep{jaderberg2017population}: An evolutionary algorithm that optimizes agent populations by iteratively mutating and selecting promising policies, facilitating diverse and robust coordination strategies.
    \item \textbf{FCP (Fictitious Co-Play)}~\citep{strouse2021collaborating}: A method that models coordination by simulating the behaviors of various partner types, enabling agents to adapt to unseen collaborators.
    \item \textbf{MEP (Maximum Entropy Population)}~\citep{zhao2023maximum}: This approach promotes diversity within agent populations by maximizing entropy, which encourages exploration of varied strategies for better coordination.
    \item \textbf{COLE (Cooperative Learning)}~\citep{li2023cooperative}: An algorithm designed to enhance cooperative behavior between agents by explicitly learning to predict and adapt to partners’ actions.
\end{itemize}

These baselines were selected due to their relevance and proven success in multi-agent coordination scenarios. The pretrained baseline models were obtained from the ProAgent GitHub repository\footnote{\url{https://github.com/PKU-Alignment/ProAgent} (MIT License)}.

We also evaluate CausalPlan in collaboration with a human policy collected via behavior learning, available at the COLE platform\footnote{\url{https://github.com/liyang619/COLE-Platform}}.

\subsection{Additional experiment results}
\label{sect A Additional Experiments}

\subsubsection{Details of AI player evaluation}
\label{sect A Human-AI Evaluation}
Tab.~\ref{tab:table A1} presents a comprehensive comparison of the performance of various backbone LLMs, both with and without CausalPlan, evaluated across multiple layouts.

Tab.~\ref{tab:table A2} provides an in-depth comparison between baseline agents and our proposed CausalPlan method using Llama-70B, across different layouts.

\begin{table}[h]
\centering
\scriptsize 
\caption{Performance of different backbones with and without CausalPlan across various layouts (this is the detailed version of Fig.~\ref{fig:backbone_compare}). The reported results, including mean and variance, are obtained from 3 different seeds, with each seed running for 400 timesteps. In these experiments, we use the small LLM agent as Player 1, allowing it to collaborate with all other baselines as described in Sect.~\ref{sect 5.1}, and report the average and variance of the outcomes. The last column reports the average improvement across backbones, and the last row reports the average improvement across layouts in \%. The result with the highest improvement is highlighted in \textbf{bold}, while the second highest is \underline{underscored}.}
\renewcommand{\arraystretch}{1.5}
\begin{tabular}{l | c | c c c c c | c}
\hline
\multirow{2}{*}{\textbf{Backbones}} & 
\multirow{2}{*}{\shortstack{\textbf{With} \\ \textbf{CausalPlan}}} & 
\multicolumn{5}{c|}{\textbf{Layouts}} & 
\multirow{2}{*}{\shortstack{\textbf{Avg.} \\ \textbf{Improv.} \\ \textbf{(\%)}}} \\ \cline{3-7}

& & \textbf{CR} & \textbf{AA} & \textbf{COR} & \textbf{FC} & \textbf{CC} & \\ \hline
\multirow{2}{*}{\textbf{Gemma-7B}} & \textbf{\(\times\)} & 121.3 $\pm$ 16.2 & 88.0 $\pm$ 32.7 & 78.7 $\pm$ 8.3 & 17.3 $\pm$ 6.1 & 73.3 $\pm$ 10.1 & \multirow{2}{*}{12.82} \\ \cline{2-7}
& \textbf{\checkmark}  & 141.3 $\pm$ 6.1  & 122.7 $\pm$ 12.9 & 82.7 $\pm$ 30.5 & 17.3 $\pm$ 9.2 & 78.7 $\pm$ 14.0 & \\ \hline
\multirow{2}{*}{\textbf{Llama-8B}} & \textbf{\(\times\)} & 110.7 $\pm$ 12.8 & 163.4 $\pm$ 3.3 & 80.0 $\pm$ 41.7 & 9.3 $\pm$ 2.3 & 84.0 $\pm$ 20.8 & \multirow{2}{*}{13.90} \\ \cline{2-7}
& \textbf{\checkmark}  & 150.7 $\pm$ 2.3  & 182.2 $\pm$ 18.3 & 77.3 $\pm$ 14.0 & 16.0 $\pm$ 4.0 & 90.7 $\pm$ 2.3 & \\ \hline
\multirow{2}{*}{\textbf{Qwen-14B}} & \textbf{\(\times\)} & 117.3 $\pm$ 4.6  & 224.0 $\pm$ 22.6 & 76.0 $\pm$ 17.4 & 16.0 $\pm$ 4.0 & 48.0 $\pm$ 22.6 & \multirow{2}{*}{\textbf{29.04}} \\ \cline{2-7}
& \textbf{\checkmark}  & 162.6 $\pm$ 9.2  & 232.0 $\pm$ 31.7 & 121.3 $\pm$ 16.6 & 17.3 $\pm$ 12.8 & 93.3 $\pm$ 22.7 & \\ \hline
\multirow{2}{*}{\textbf{Llama-70B}} & \textbf{\(\times\)} & 144.0 $\pm$ 18.3 & 248.0 $\pm$ 22.7 & 125.3 $\pm$ 10.0 & 34.7 $\pm$ 14.0 & 89.3 $\pm$ 32.3 & \multirow{2}{*}{\underline{22.42}} \\ \cline{2-7}
& \textbf{\checkmark}  & 178.7 $\pm$ 2.3  & 266.7 $\pm$ 16.7 & 157.3 $\pm$ 2.3 & 38.7 $\pm$ 16.2 & 112.0 $\pm$ 6.9 & \\ \hline
\textbf{Avg. Improv. (\%)} & -- & \textbf{20.83} & 18.80 & \underline{19.13} & 4.87 & 9.55 & -- \\ \hline
\textbf{Oracle GPT} & -- & 194.2 $\pm$ 10.5 & 229.8 $\pm$ 21.9 & 183.0 $\pm$ 31.7 & 31.0 $\pm$ 33.9 & 128.5 $\pm$ 28.1 & -- \\ \hline
\end{tabular}
\label{tab:table A1}
\end{table}

\begin{table}[h]
\caption{Performance comparison between baseline agents and CausalPlan (Ours) across layouts using Llama-70B (this is the detailed version of Tab.~\ref{tab:table 1}). Results (mean ± variance) are averaged over 3 seeds (400 timesteps each). The first row per layout corresponds to our agent as Player 0, the second to Player 1. Best and second-best results are in \textbf{bold} and \underline{underlined}, respectively.
}
\label{tab:table A2}
\centering
\small   
\renewcommand{\arraystretch}{1.2}  
\begin{tabular}{l|c|c|c|c|c|c}
\hline
\multirow{2}{*}{\textbf{Layout}} & \multicolumn{5}{c|}{\textbf{Baseline AI Agents}} & \multirow{2}{*}{\textbf{\shortstack{CausalPlan \\ (Ours)}}} \\ \cline{2-6}
                                  & \textbf{SP} & \textbf{PBT} & \textbf{FCP} & \textbf{MEP} & \textbf{COLE} & \\ \hline
\multirow{2}{*}{\textbf{CR}}              & 160.0 $\pm$ 4.0 & 165.3 $\pm$ 1.7 & \textbf{194.6 $\pm$ 10.0} & \underline{177.3 $\pm$ 22.0} & 164.0 $\pm$ 6.9 & 166.7 $\pm$ 6.1 \\ \cline{2-7}
                                  & 164.0 $\pm$ 16.0 & 170.7 $\pm$ 8.3 & \textbf{193.3 $\pm$ 10.1} & 178.7 $\pm$ 10.1 & 142.7 $\pm$ 18.0 & \underline{178.7 $\pm$ 2.3} \\ \hline
\multirow{2}{*}{\textbf{AA}}               & 173.3 $\pm$ 22.0 & 185.3 $\pm$ 12.8 & 181.3 $\pm$ 14.0 & 153.3 $\pm$ 2.3 & \underline{197.3 $\pm$ 14.0} & \textbf{250.7 $\pm$ 16.1} \\ \cline{2-7}
                                  & \underline{194.7 $\pm$ 12.9} & 150.7 $\pm$ 18.0 & 172.0 $\pm$ 16.0 & 181.3 $\pm$ 9.2 & 173.3 $\pm$ 16.2 & \textbf{266.7 $\pm$ 16.7} \\ \hline
\multirow{2}{*}{\textbf{COR}}              & 106.7 $\pm$ 12.8 & 138.7 $\pm$ 12.2 & 138.7 $\pm$ 2.3 & \textbf{166.7 $\pm$ 8.3} & 154.7 $\pm$ 2.3 & \underline{156.0 $\pm$ 4.0} \\ \cline{2-7}
                                  & 134.7 $\pm$ 9.2 & 140.0 $\pm$ 8.0 & 122.7 $\pm$ 10.1 & \underline{154.7 $\pm$ 6.1} & 152.0 $\pm$ 6.9 & \textbf{157.3 $\pm$ 2.3}\\ \hline
\multirow{2}{*}{\textbf{FC}}               & 10.7 $\pm$ 4.6  & 20.0 $\pm$ 14.4  & \underline{57.3 $\pm$ 6.1}  & 22.7 $\pm$ 4.6  & 41.3 $\pm$ 10.0 & \textbf{69.1 $\pm$ 13.6} \\ \cline{2-7}
                                  & 25.3 $\pm$ 4.6 & \textbf{61.3 $\pm$ 6.1} & 26.7 $\pm$ 8.3 & 38.0 $\pm$ 6.1 &
                                  \underline{48.0 $\pm$ 4.0} & 38.7 $\pm$ 16.1 \\ \hline
\multirow{2}{*}{\textbf{CC}}               & 62.7 $\pm$ 12.2 & 56.0 $\pm$ 8.0 & 64.0 $\pm$ 8.0 & 33.3 $\pm$ 22.0 & \underline{96.0 $\pm$ 4.0} & \textbf{113.3 $\pm$ 8.3} \\ \cline{2-7}
                                  & 50.7 $\pm$ 6.1 & 48.0 $\pm$ 20.0 & 62.7 $\pm$ 12.9 & 66.7 $\pm$ 10.1 & \underline{85.3 $\pm$ 16.2} & \textbf{112.0 $\pm$ 6.9} \\ \hline
\end{tabular}
\end{table}

\subsubsection{Effect of hyperparameter $\gamma$}
\label{sect A Effect of gamma Hyperparameter}

\begin{figure}[h]
\begin{center}
    \includegraphics[width=0.65\textwidth]{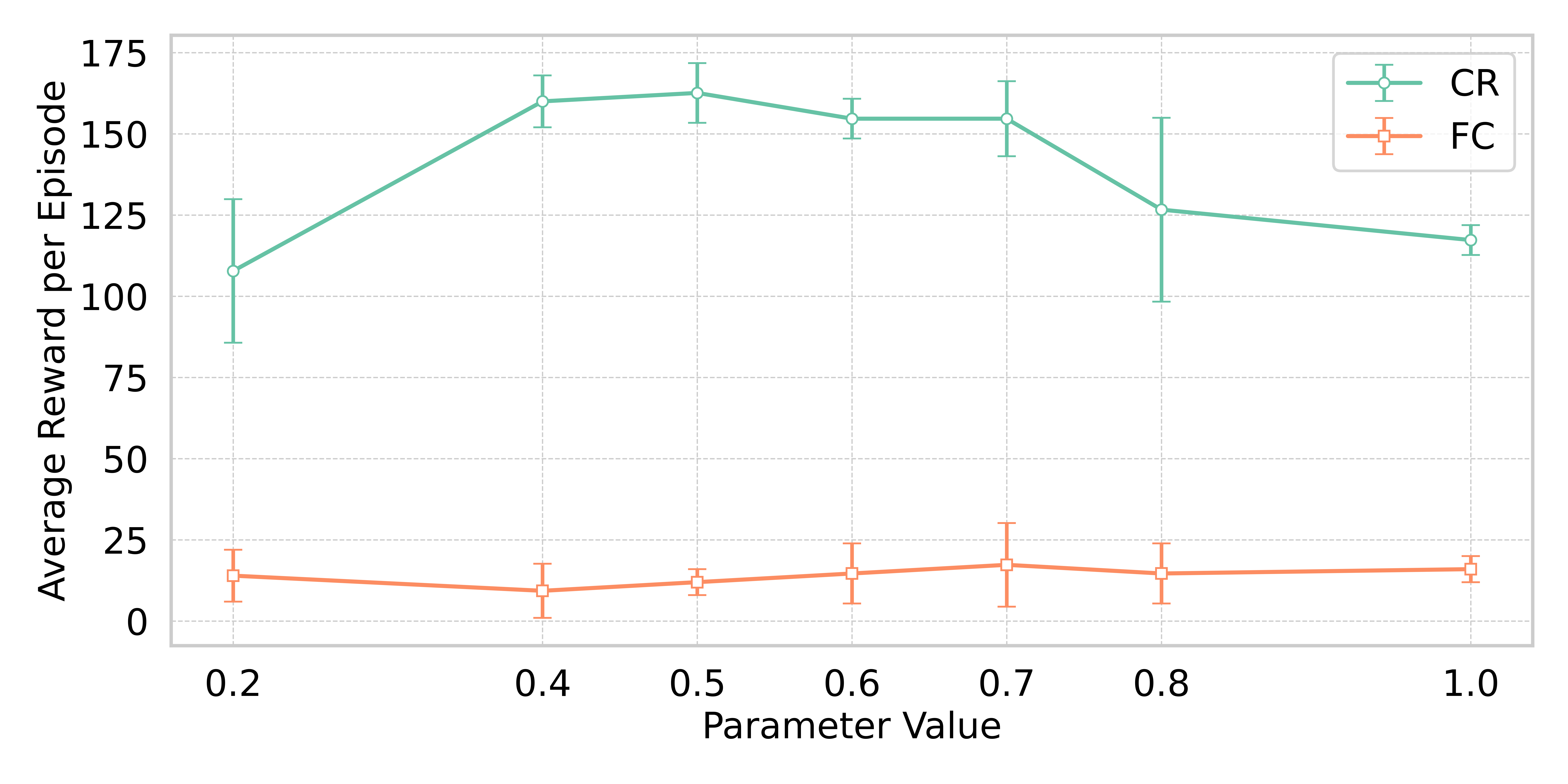}  
\end{center}
\caption{Experiments showing the impact of tuning the hyperparameter \(\gamma\) conducted using Qwen-14B on CR and FC layouts. The results, including mean and variance, are averaged over three different seeds. The optimal value of \(\gamma\) typically lies within the range of 0.4-0.8, emphasizing the importance of balancing between the belief of the LLMs and the prior causal knowledge.}
\label{fig:gamma_tuning}  
\end{figure}

In our framework, the hyperparameter $\gamma$ in Eq. 10 controls the balance between the agent's belief and the causal knowledge. To investigate the effect of varying $\gamma$, we conducted an experiment on two layouts, CR and FC, using Qwen-14B as the backbone LLM. As shown in Fig.~\ref{fig:gamma_tuning}, the optimal value for $\gamma$ lies within the range of 0.5 to 0.7. In both cases, when $\gamma$ is set to 0.2, indicating a greater reliance on causal knowledge than on the agent's own knowledge, or when $\gamma$ is set to 1, fully trusting the agent, the performance degraded.
Refer to Fig.~\ref{fig:gamma_tuning} for the experimental results and Tab.~\ref{tab:gamma_values} for the $\gamma$ values used for each LLM agent across different layouts. Due to limited computational resources, tuning was only performed on layouts where CausalPlan initially underperformed with $\gamma=0.5$. We believe that further tuning of this hyperparameter would likely lead to improved performance.

\subsubsection {Effect of different data collection policy}
\label{sect A Effect of Different Data Collection Agents}

\begin{table}[h]
\caption{Ablation studies on using different agents to collect data for buffer $B$ conducted on CR layout with Llama-8B as backbone. The results, including mean and variance, are obtained from 3 different seeds. ''Llama-8B'' and ''MEP'' refer to using Llama-8B or MEP to generate data.}
\label{tab:table A3}
\centering
\small
\renewcommand{\arraystretch}{2}  
\setlength{\tabcolsep}{2pt}  
\begin{tabular}{l|c|c|c|c|c|c}
\hline
\multirow{2}{*}{\textbf{Methods}} & \multicolumn{5}{c|}{\textbf{Baseline AI Agents}} & \multirow{2}{*}{\shortstack{\textbf{Average} \\ \textbf{Results}}} \\ \cline{2-6}

                                 & SP & PBT & FCP & MEP & COLE & \\ \hline
\textbf{Llama-8B}       & 106.7 $\pm$ 41.6 & 86.7 $\pm$ 75.1 & 166.7 $\pm$ 41.6 & 126.7 $\pm$ 30.6 & 140.0 $\pm$ 0.0 & 125.3 $\pm$ 30.7  \\ \hline
\textbf{MEP}             & 126.7 $\pm$ 30.6 & 133.3 $\pm$ 30.5 & 160.0 $\pm$ 40.0 & 166.7 $\pm$ 41.6 & 166.7 $\pm$ 23.1 & \textbf{150.7 $\pm$ 2.3} \\ \hline
\end{tabular}
\end{table}

Tab.~\ref{tab:table A3} presents an ablation study comparing the effects of using different agents—Llama-8B and MEP—for data collection in buffer \( B \), when interacting with the environment to collect data for 200k steps. We hypothesize that using MEP for data collection would yield better results, given that it is a pretrained agent specialized for the task. Nevertheless, even when using data collected by Llama-8B, incorporating causal knowledge still provides a performance gain compared to not using causal knowledge at all. The results show that MEP consistently outperforms Llama-8B across all baseline AI agents, achieving a higher average score of 150.7 (±2.3) compared to 125.3 (±30.7) for Llama-8B. This underscores the importance of utilizing a stronger agent to generate high-quality training data for causal reasoning. Importantly, even when using data from Llama-8B, causal knowledge improves performance relative to the absence of causal guidance, where the average score drops to 110.7 (±12.8) as reported in Appx. Tab.~\ref{tab:table A1}. We hypothesize that the performance gain observed when using data from Llama-8B arises from its ability to consult not only the current deterministic action selection but also similar past scenarios through the incorporation of causal knowledge.

\subsubsection {Heatmap of learned causal matrix $\mathcal{M}$ analysis}
\label{sect A Heatmap of Learned Causal Matrix}
\begin{figure}[H]
\begin{center}
    \includegraphics[width=1\textwidth]{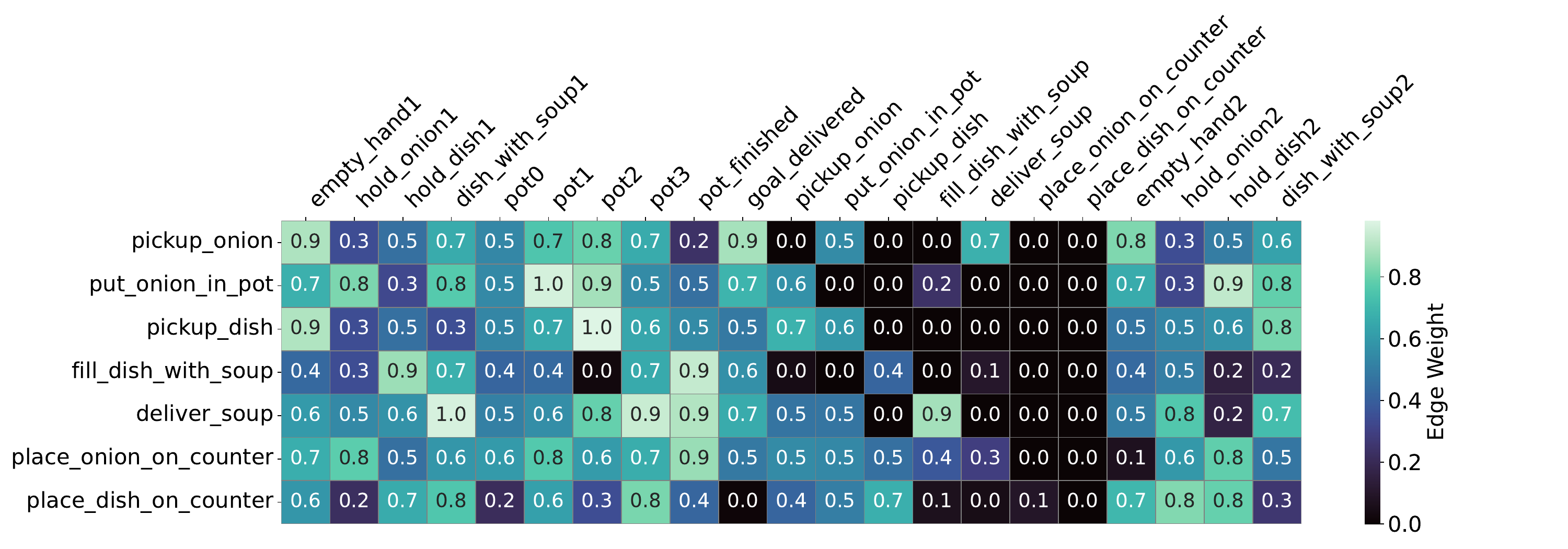}  
\end{center}
\caption{Heatmap of causal graph edge weights obtained from data collected using MEP in the CR layout. The plot illustrates the influence of state features (x-axis) on agent actions (y-axis).}
\label{fig:heatmap_mep}  
\end{figure}

\begin{figure}[H]
\begin{center}
    \includegraphics[width=1\textwidth]{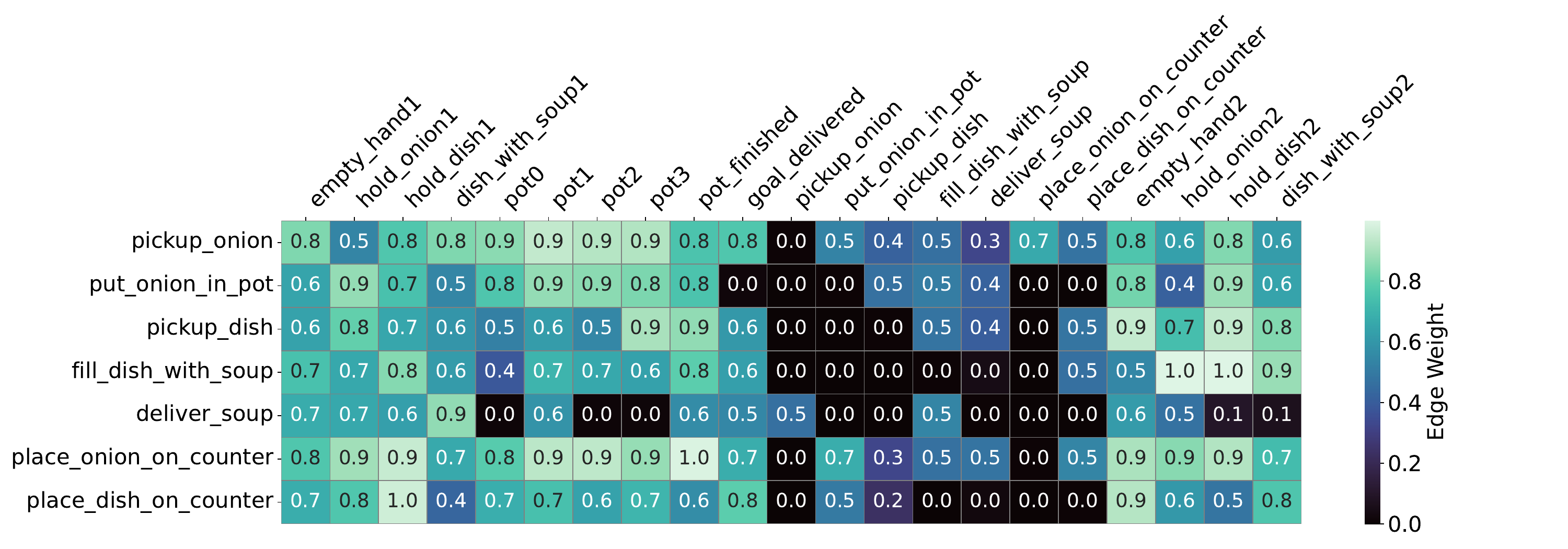}  
\end{center}
\caption{Heatmap of causal graph edge weights obtained from data collected using the Llama-8B backbone in the CR layout. The plot illustrates the influence of state features (x-axis) on agent actions (y-axis).}
\label{fig:heatmap_llama}  
\end{figure}

In Fig.~\ref{fig:heatmap_mep} and Fig.~\ref{fig:heatmap_llama}, we present the causal matrices \(\mathcal{M}\) derived from data collected by MEP and Llama-8B, respectively. The inference results using these matrices are detailed in Appx.~\ref{sect A Effect of Different Data Collection Agents}. To obtain each matrix, the respective agent interacts with the environment for 200,000 steps to gather data, followed by training the SCA model for 500,000 steps on the collected dataset. While both matrices share similarities in many key edges—for example, from \texttt{empty\_hand1} to \texttt{pickup\_onion} (edge weights of 0.9 for MEP and 0.8 for Llama-8B) and from \texttt{pot\_finished} to \texttt{fill\_dish\_with\_soup} (0.9 for MEP and 0.8 for Llama-8B) (see Appx.~\ref{sect A States and actions factorization in each environment} for feature descriptions)—there are important differences that likely contribute to performance variations. For instance, the edge from \texttt{pickup\_onion} to \texttt{put\_onion\_in\_pot} has a weight of 0.6 when using MEP-collected data but is absent (weight 0) with Llama-collected data. Similarly, the transition from \texttt{deliver\_soup} to \texttt{pickup\_onion} appears with a weight of 0.7 in the MEP matrix but is missing in the Llama-8B matrix. These differences highlight how the choice of data collection agent influences the learned causal structure, which in turn can impact the effectiveness of downstream inference and control.

Additionally, one may observe that both heatmaps contain several edges that are difficult to interpret, especially those originating from the state of the other agent toward the current action. These edges may carry meaning for the agent but appear unintelligible to humans, or they may be irrelevant. However, these unexpected edges have minimal impact on the inference process, provided the LLM agent does not sample the corresponding actions, thereby eliminating the need to re-calculate the final associated sampling probabilities. This highlights the importance of the general knowledge embedded within the LLM agent, which helps partially eliminate irrelevant edges and leaves only those ambiguities that require causal reasoning. We hypothesize that more advanced causal discovery techniques could further improve the quality of the learned causal graphs by eliminating spurious edges. A simpler alternative might involve hyperparameter tuning of a threshold, where edges with probabilities below this threshold are removed entirely, or collecting more data. We leave these explorations for future work.

\subsubsection{Time Efficiency Analysis}
\label{sect:A_Time_Efficiency_Analysis}

Learning the causal graph—such as in the CR environment, which involves 21 parent nodes and 7 child nodes—requires approximately 3 hours of training. However, this is a one-time offline process that can be reused across all backbone models, making its cost negligible in the overall training pipeline.

The actual runtime during planning varies depending on the backbone model used. Using NVIDIA h100 GPUs (detailed in Appx.~\ref{sect A CausalPlan Implementation}, we observe the following runtimes for 400 timesteps:

\begin{itemize}
    \item Gemma-7B and Llama-8B: Approximately 5 minutes without CausalPlan, and around 15 minutes with CausalPlan.
    \item Qwen-14B: Roughly 16 minutes without CausalPlan, and 41 minutes with CausalPlan.
    \item Llama-70B: About 40 minutes without CausalPlan, and approximately 68 minutes with CausalPlan.
\end{itemize}

These results highlight the additional computational cost introduced by causal reasoning. However, the overhead remains reasonable given the observed improvements in policy quality.

\subsection{Hyperparameters}
\label{sect A Hyperparameters}
\subsubsection{Causality and LLMs hyperparameters}
\label{sect A Causality and LLMs hyperparameters}

\textbf{SCA Model}

\begin{table}[h!]
\centering
\caption{Hyperparameters related to SCA Model}
\begin{tabular}{llp{7cm}}
\toprule
\textbf{Parameter} & \textbf{Value} & \textbf{Description} \\
\midrule
$N$ & 200{,}000 & Timesteps used to collect data for buffer $B$. \\
$T$ & 400 & Horizon of each episode. \\
$f_i$ network architecture & MLP & Four hidden layers with dimensions 64, 256, 256, 64; ReLU activations; sigmoid output. \\
Optimizer & Adam & Optimization algorithm used to train $\delta$ and $\eta$. \\
Learning rate & 3e-4 & Step size for gradient updates for $\delta$ and $\eta$. \\
Regularization $\lambda$ & 1e-7 & Regularization strength for parameter estimation. \\
Iterations & 500{,}000 & Number of training iterations for $\delta$ and $\eta$. \\
\bottomrule
\end{tabular}
\end{table}

\vspace{0.5cm}

\textbf{LLMs Agent (Build on top of ProAgent framework~\citep{zhang2024proagent}}

\begin{table}[h!]
\centering
\caption{Hyperparameters related to LLMs}
\begin{tabular}{lp{2cm}p{7cm}}
\toprule
\textbf{Parameter} & \textbf{Value} & \textbf{Description} \\
\midrule

Model sizes & Gemma-7B, Qwen-14B, Llama-8B, Llama-70B & Language model sizes and architectures \\

Temperature & 1.0 & Controls randomness; higher values encourage diverse samples \\
Max new tokens & 256 & Maximum number of generated tokens per output \\
Top-k sampling & 50 & Number of top tokens considered in sampling \\
Top-p sampling & 0.9 & Nucleus sampling threshold (alternative to top-k) \\
Sampling method & Enabled & Sampling is enabled (do\_sample=true) \\
retrival\_method & recent\_k & Parameter of ProAgent framework to retrieve recent history dialogue \\
K & 1 & Parameter of ProAgent framework, the number of history dialogue (default value is 0 or 1) \\
\bottomrule
\end{tabular}
\end{table}

\textbf{$\gamma$ value in Eq.~\ref{eq:6} for each layouts}

\begin{table}[h]
\centering
\caption{\textbf{$\gamma$ value in Eq.~\ref{eq:6} for each layout and language model}}
\label{tab:gamma_values}
\begin{tabular}{lcccc}
\toprule
\textbf{Layout} & \textbf{Gemma-7B} & \textbf{Qwen-14B} & \textbf{Llama-8B} & \textbf{Llama-70B} \\
\midrule
CR & 0.5 & 0.5 & 0.5 & 0.5 \\
AA & 0.5 & 0.5 & 0.5 & 0.5 \\
COR & 0.5 & 0.5 & 0.5 & 0.5 \\
FC & 0.6 & 0.7 & 0.4 & 0.5 \\
CC & 0.5 & 0.5 & 0.5 & 0.5 \\
\bottomrule
\end{tabular}
\end{table}

\subsubsection{States and actions factorization in each environment}
\label{sect A States and actions factorization in each environment}

States and actions factorization used in CR layouts are available in Tab.~\ref{tab:CR_factors} and for other other layouts are included in Tab.~\ref{tab:Other_factors}.

\begin{table}[h]
\centering
\caption{Factorized States and Actions for CR Layout with Descriptions}
\begin{tabular}{ll}
\toprule
\textbf{Feature} & \textbf{Description} \\
\midrule
\texttt{empty\_hand1}           & Controlling agent is not holding any object \\
\texttt{hold\_onion1}           & Controlling agent is holding an onion \\
\texttt{hold\_dish1}            & Controlling agent is holding an empty dish \\
\texttt{dish\_with\_soup1}      & Controlling agent is holding a dish filled with soup \\
\midrule
\texttt{pot0}                   & Pot contains 0 onions (empty) \\
\texttt{pot1}                   & Pot contains 1 onion \\
\texttt{pot2}                   & Pot contains 2 onions \\
\texttt{pot3}                   & Pot contains 3 onions (ready to cook) \\
\texttt{pot\_finished}          & Pot has finished cooking and soup is ready \\
\midrule
\texttt{goal\_delivered}        & A soup has been successfully delivered to the goal \\
\midrule
\texttt{pickup\_onion}             & Action: controlling agent picks up an onion \\
\texttt{put\_onion\_in\_pot}      & Action: controlling agent places an onion into a pot \\
\texttt{pickup\_dish}              & Action: controlling agent picks up an empty dish \\
\texttt{fill\_dish\_with\_soup}   & Action: controlling agent fills a dish with soup from a finished pot \\
\texttt{deliver\_soup}             & Action: controlling agent delivers a soup to the goal \\
\texttt{place\_onion\_on\_counter} & Action: controlling agent places an onion on the counter \\
\texttt{place\_dish\_on\_counter}  & Action: controlling agent places a dish on the counter \\
\midrule
\texttt{empty\_hand2}           & Other agent is not holding any object \\
\texttt{hold\_onion2}           & Other agent is holding an onion \\
\texttt{hold\_dish2}            & Other agent is holding an empty dish \\
\texttt{dish\_with\_soup2}      & Other agent is holding a dish filled with soup \\
\bottomrule
\end{tabular}
\label{tab:CR_factors}
\end{table}

\begin{table}[h]
\centering
\caption{Factorized states and actions for other layouts and their descriptions}
\begin{tabular}{ll}
\toprule
\textbf{Feature} & \textbf{Description} \\
\midrule
\texttt{empty\_hand1}      & Controlling agent is not holding any object \\
\texttt{hold\_onion1}      & Controlling agent is holding an onion \\
\texttt{hold\_dish1}       & Controlling agent is holding an empty dish \\
\texttt{dish\_with\_soup1} & Controlling agent is holding a dish filled with soup \\
\midrule
\texttt{pot0\_0}           & Pot 0 contains 0 onions (empty) \\
\texttt{pot1\_0}           & Pot 0 contains 1 onion \\
\texttt{pot2\_0}           & Pot 0 contains 2 onions \\
\texttt{pot3\_0}           & Pot 0 contains 3 onions (ready to cook) \\
\texttt{pot\_finished\_0}    & Pot 0 has finished cooking and soup is ready \\
\midrule
\texttt{pot0\_1}        & Pot 1 contains 0 onions (empty) \\
\texttt{pot1\_1}        & Pot 1 contains 1 onion \\
\texttt{pot2\_1}        & Pot 1 contains 2 onions \\
\texttt{pot3\_1}        & Pot 1 contains 3 onions (ready to cook) \\
\texttt{pot\_finished\_1} & Pot 1 has finished cooking and soup is ready \\
\midrule
\texttt{goal\_delivered}  & A soup has been successfully delivered to the goal \\
\midrule
\texttt{pickup\_onion}             & Action: controlling agent picks up an onion \\
\texttt{put\_onion\_in\_pot}      & Action: controlling agent places an onion into a pot \\
\texttt{pickup\_dish}              & Action: controlling agent picks up an empty dish \\
\texttt{fill\_dish\_with\_soup}   & Action: controlling agent fills a dish with soup from a finished pot \\
\texttt{deliver\_soup}             & Action: controlling agent delivers a soup to the goal \\
\texttt{place\_onion\_on\_counter} & Action: controlling agent places an onion on the counter \\
\texttt{place\_dish\_on\_counter}  & Action: controlling agent places a dish on the counter \\
\midrule
\texttt{empty\_hand2}      & Other agent is not holding any object \\
\texttt{hold\_onion2}      & Other agent is holding an onion \\
\texttt{hold\_dish2}       & Other agent is holding an empty dish \\
\texttt{dish\_with\_soup2} & Other agent is holding a dish filled with soup \\
\bottomrule
\end{tabular}
\label{tab:Other_factors}
\end{table}

\section{Discussion of broader impacts}
\label{sect Broader Impacts}
This work represents an important foundational step toward integrating causal reasoning into multi-agent planning with large language models (LLMs). Our causality-driven framework aims to improve the safety, efficiency, and interpretability of collaborative AI systems by enabling agents to better understand the consequences of their states and actions. Although primarily exploratory and not yet intended for real-world deployment, the results demonstrate promising potential for advancing multi-agent coordination.

At this stage, we do not expect any direct negative societal impacts, as the framework requires further development and validation before practical use. Nevertheless, as autonomous multi-agent systems mature, concerns related to fairness, reliability, misuse, and broader ethical implications will become increasingly important. Addressing these challenges through responsible design, transparency, and rigorous evaluation will be critical to ensure the safe and trustworthy deployment of such systems in the future.

%% file: main.bbl
\begin{thebibliography}{40}
\providecommand{\natexlab}[1]{#1}
\providecommand{\url}[1]{\texttt{#1}}
\expandafter\ifx\csname urlstyle\endcsname\relax
  \providecommand{\doi}[1]{doi: #1}\else
  \providecommand{\doi}{doi: \begingroup \urlstyle{rm}\Url}\fi

\bibitem[Achiam et~al.(2023)Achiam, Adler, Agarwal, Ahmad, Akkaya, Aleman, Almeida, Altenschmidt, Altman, Anadkat, et~al.]{openai2024gpt4technicalreport}
Josh Achiam, Steven Adler, Sandhini Agarwal, Lama Ahmad, Ilge Akkaya, Florencia~Leoni Aleman, Diogo Almeida, Janko Altenschmidt, Sam Altman, Shyamal Anadkat, et~al.
\newblock {GPT}-4 {T}echnical {R}eport.
\newblock \emph{arXiv preprint arXiv:2303.08774}, 2023.

\bibitem[Ahn et~al.(2022)Ahn, Brohan, Brown, Chebotar, Cortes, David, Finn, Fu, Gopalakrishnan, Hausman, et~al.]{ahn2022can}
Michael Ahn, Anthony Brohan, Noah Brown, Yevgen Chebotar, Omar Cortes, Byron David, Chelsea Finn, Chuyuan Fu, Keerthana Gopalakrishnan, Karol Hausman, et~al.
\newblock Do as {I} can, not as {I} say: grounding language in robotic affordances.
\newblock \emph{arXiv preprint arXiv:2204.01691}, 2022.

\bibitem[Carroll et~al.(2019)Carroll, Shah, Ho, Griffiths, Seshia, Abbeel, and Dragan]{carroll2019utility}
Micah Carroll, Rohin Shah, Mark~K Ho, Thomas~L Griffiths, Sanjit~A Seshia, Pieter Abbeel, and Anca Dragan.
\newblock On the utility of learning about humans for human-{AI} coordination.
\newblock In \emph{Advances in Neural Information Processing Systems (NeurIPS)}, pages 5174--5185, 2019.

\bibitem[Cohere(2024)]{cohere2024commandr}
Cohere.
\newblock The command {R} model (details and application).
\newblock \url{https://docs.cohere.com/v2/docs/command-r}, 2024.
\newblock Accessed: 2025-05-12.

\bibitem[Corcoll and Vicente(2020)]{corcoll2020disentangling}
Oriol Corcoll and Raul Vicente.
\newblock Disentangling causal effects for hierarchical reinforcement learning.
\newblock \emph{arXiv preprint arXiv:2010.01351}, 2020.

\bibitem[Cross et~al.(2024)Cross, Xiang, Bhatia, Yamins, and Haber]{cross2024hypothetical}
Logan Cross, Violet Xiang, Agam Bhatia, Daniel~LK Yamins, and Nick Haber.
\newblock Hypothetical minds: scaffolding theory of mind for multi-agent tasks with large language models.
\newblock In \emph{NeurIPS 2024 Workshop on Open-World Agents}, 2024.

\bibitem[Du et~al.(2024)Du, Ye, Zhang, Yang, Chen, and Wang]{du2024situation}
Xiao Du, Yutong Ye, Pengyu Zhang, Yaning Yang, Mingsong Chen, and Ting Wang.
\newblock Situation-dependent causal influence-based cooperative multi-agent reinforcement learning.
\newblock In \emph{Proceedings of the AAAI Conference on Artificial Intelligence}, pages 17362--17370, 2024.

\bibitem[Gao et~al.(2023{\natexlab{a}})Gao, Lan, Lu, Mao, Piao, Wang, Jin, and Li]{gao2023s3}
Chen Gao, Xiaochong Lan, Zhihong Lu, Jinzhu Mao, Jinghua Piao, Huandong Wang, Depeng Jin, and Yong Li.
\newblock S3: Social-network simulation system with large language model-empowered agents.
\newblock \emph{arXiv preprint arXiv:2307.14984}, 2023{\natexlab{a}}.

\bibitem[Gao et~al.(2023{\natexlab{b}})Gao, Ding, Qin, and Liu]{gao2023chatgpt}
Jinglong Gao, Xiao Ding, Bing Qin, and Ting Liu.
\newblock Is {C}hat{GPT} a good causal reasoner? {A} comprehensive evaluation.
\newblock In \emph{Findings of the Association for Computational Linguistics: EMNLP 2023}, pages 11111--11126, 2023{\natexlab{b}}.

\bibitem[Guo et~al.(2025)Guo, Yang, Zhang, Song, Zhang, Xu, Zhu, Ma, Wang, Bi, et~al.]{deepseekai2025deepseekr1incentivizingreasoningcapability}
Daya Guo, Dejian Yang, Haowei Zhang, Junxiao Song, Ruoyu Zhang, Runxin Xu, Qihao Zhu, Shirong Ma, Peiyi Wang, Xiao Bi, et~al.
\newblock Deepseek-{R}1: incentivizing reasoning capability in {LLM}s via reinforcement learning.
\newblock \emph{arXiv preprint arXiv:2501.12948}, 2025.

\bibitem[Hu et~al.(2020)Hu, Lerer, Peysakhovich, and Foerster]{hu2020other}
Hengyuan Hu, Adam Lerer, Alex Peysakhovich, and Jakob Foerster.
\newblock "{O}ther-play" for zero-shot coordination.
\newblock In \emph{Proceedings of the International Conference on Machine Learning (ICML)}, pages 4399--4410, 2020.

\bibitem[Jaderberg et~al.(2017)Jaderberg, Dalibard, Osindero, Czarnecki, Donahue, Razavi, Vinyals, Green, Dunning, Simonyan, Fernando, and Kavukcuoglu]{jaderberg2017population}
Max Jaderberg, Valentin Dalibard, Simon Osindero, Wojciech~M Czarnecki, Jeff Donahue, Ali Razavi, Oriol Vinyals, Tim Green, Iain Dunning, Karen Simonyan, Chrisantha Fernando, and Koray Kavukcuoglu.
\newblock Population based training of neural networks.
\newblock \emph{arXiv preprint arXiv:1711.09846}, 2017.

\bibitem[Jaques et~al.(2019)Jaques, Lazaridou, Hughes, Gulcehre, Ortega, Strouse, Leibo, and De~Freitas]{jaques2019social}
Natasha Jaques, Angeliki Lazaridou, Edward Hughes, Caglar Gulcehre, Pedro Ortega, DJ~Strouse, Joel~Z Leibo, and Nando De~Freitas.
\newblock Social influence as intrinsic motivation for multi-agent deep reinforcement learning.
\newblock In \emph{Proceedings of the International Conference on Machine Learning (ICML)}, pages 3040--3049, 2019.

\bibitem[Ke et~al.(2019)Ke, Bilaniuk, Goyal, Bauer, Larochelle, Sch{\"o}lkopf, Mozer, Pal, and Bengio]{ke2019learning}
Nan~Rosemary Ke, Olexa Bilaniuk, Anirudh Goyal, Stefan Bauer, Hugo Larochelle, Bernhard Sch{\"o}lkopf, Michael~C Mozer, Chris Pal, and Yoshua Bengio.
\newblock Learning neural causal models from unknown interventions.
\newblock \emph{arXiv preprint arXiv:1910.01075}, 2019.

\bibitem[Kojima et~al.(2022)Kojima, Gu, Reid, Matsuo, and Iwasawa]{kojima2022large}
Takeshi Kojima, Shixiang~Shane Gu, Machel Reid, Yutaka Matsuo, and Yusuke Iwasawa.
\newblock Large language models are zero-shot reasoners.
\newblock In \emph{Advances in Neural Information Processing Systems (NeurIPS)}, pages 22199--22213, 2022.

\bibitem[Legg and Hutter(2007)]{legg2007universal}
Shane Legg and Marcus Hutter.
\newblock Universal intelligence: A definition of machine intelligence.
\newblock \emph{Minds and {M}achines}, 17:\penalty0 391--444, 2007.

\bibitem[Li et~al.(2023)Li, Zhang, Sun, Du, Wen, Wang, and Pan]{li2023cooperative}
Yang Li, Shao Zhang, Jichen Sun, Yali Du, Ying Wen, Xinbing Wang, and Wei Pan.
\newblock Cooperative open-ended learning framework for zero-shot coordination.
\newblock In \emph{Proceedings of the International Conference on Machine Learning (ICML)}, pages 20470--20484, 2023.

\bibitem[Park et~al.(2023)Park, O'Brien, Cai, Morris, Liang, and Bernstein]{park2023generative}
Joon~Sung Park, Joseph O'Brien, Carrie~Jun Cai, Meredith~Ringel Morris, Percy Liang, and Michael~S Bernstein.
\newblock Generative agents: interactive simulacra of human behavior.
\newblock In \emph{Proceedings of the {A}nnual {ACM} {S}ymposium on {U}ser {I}nterface {S}oftware and {T}echnology}, pages 1--22, 2023.

\bibitem[Pearl(2009)]{pearl2009causality}
Judea Pearl.
\newblock \emph{Causality}.
\newblock Cambridge University Press, 2009.

\bibitem[Peng et~al.(2022)Peng, Hu, Zhang, Tang, Guo, Yi, Chen, Zhang, Du, Li, Guo, and Chen]{peng2022causality}
Shaohui Peng, Xing Hu, Rui Zhang, Ke~Tang, Jiaming Guo, Qi~Yi, Ruizhi Chen, Xishan Zhang, Zidong Du, Ling Li, Qi~Guo, and Yunji Chen.
\newblock Causality-driven hierarchical structure discovery for reinforcement learning.
\newblock In \emph{Advances in Neural Information Processing Systems (NeurIPS)}, pages 20064--20076, 2022.

\bibitem[Pitis et~al.(2020)Pitis, Creager, and Garg]{pitis2020counterfactual}
Silviu Pitis, Elliot Creager, and Animesh Garg.
\newblock Counterfactual data augmentation using locally factored dynamics.
\newblock In \emph{Advances in Neural Information Processing Systems (NeurIPS)}, pages 3976--3990, 2020.

\bibitem[Pitis et~al.(2022)Pitis, Creager, Mandlekar, and Garg]{pitis2022mocoda}
Silviu Pitis, Elliot Creager, Ajay Mandlekar, and Animesh Garg.
\newblock Mocoda: model-based counterfactual data augmentation.
\newblock In \emph{Advances in Neural Information Processing Systems (NeurIPS)}, pages 18143--18156, 2022.

\bibitem[Qian et~al.(2024)Qian, Xie, Wang, Liu, Dang, Du, Chen, Yang, Liu, and Sun]{qian2024scaling}
Chen Qian, Zihao Xie, Yifei Wang, Wei Liu, Yufan Dang, Zhuoyun Du, Weize Chen, Cheng Yang, Zhiyuan Liu, and Maosong Sun.
\newblock Scaling large-language-model-based multi-agent collaboration.
\newblock \emph{arXiv preprint arXiv:2406.07155}, 2024.

\bibitem[Qiao et~al.(2024)Qiao, Fang, Zhang, Zhu, Chen, Deng, Jiang, Xie, Huang, and Chen]{qiao2024agent}
Shuofei Qiao, Runnan Fang, Ningyu Zhang, Yuqi Zhu, Xiang Chen, Shumin Deng, Yong Jiang, Pengjun Xie, Fei Huang, and Huajun Chen.
\newblock Agent planning with world knowledge model.
\newblock In \emph{Advances in Neural Information Processing Systems (NeurIPS)}, volume~37, pages 114843--114871, 2024.

\bibitem[Seitzer et~al.(2021)Seitzer, Sch{\"o}lkopf, and Martius]{seitzer2021causal}
Maximilian Seitzer, Bernhard Sch{\"o}lkopf, and Georg Martius.
\newblock Causal influence detection for improving efficiency in reinforcement learning.
\newblock In \emph{Advances in Neural Information Processing Systems (NeurIPS)}, pages 22905--22918, 2021.

\bibitem[Shinn et~al.(2023)Shinn, Cassano, Gopinath, Narasimhan, and Yao]{shinn2023reflexion}
Noah Shinn, Federico Cassano, Ashwin Gopinath, Karthik Narasimhan, and Shunyu Yao.
\newblock Reflexion: language agents with verbal reinforcement learning.
\newblock In \emph{Advances in Neural Information Processing Systems (NeurIPS)}, pages 8634--8652, 2023.

\bibitem[Strouse et~al.(2021)Strouse, McKee, Botvinick, Hughes, and Everett]{strouse2021collaborating}
DJ~Strouse, Kevin~R McKee, Matt Botvinick, Edward Hughes, and Richard Everett.
\newblock Collaborating with humans without human data.
\newblock In \emph{Advances in Neural Information Processing Systems (NeurIPS)}, pages 14502--14515, 2021.

\bibitem[Tesauro(1994)]{tesauro1994td}
Gerald Tesauro.
\newblock {TD}-{G}ammon, a self-teaching backgammon program, achieves master-level play.
\newblock \emph{Neural Computation}, 6\penalty0 (2):\penalty0 215--219, 1994.

\bibitem[Wang et~al.(2022)Wang, Wei, Schuurmans, Le, Chi, Narang, Chowdhery, and Zhou]{wang2022self}
Xuezhi Wang, Jason Wei, Dale Schuurmans, Quoc~V Le, Ed~H Chi, Sharan Narang, Aakanksha Chowdhery, and Denny Zhou.
\newblock Self-consistency improves chain of thought reasoning in language models.
\newblock In \emph{Proceedings of the International Conference on Learning Representations (ICLR)}, 2022.

\bibitem[Wei et~al.(2022)Wei, Wang, Schuurmans, Bosma, Ichter, Xia, Chi, Le, and Zhou]{wei2022chain}
Jason Wei, Xuezhi Wang, Dale Schuurmans, Maarten Bosma, Brian Ichter, Fei Xia, Ed~H Chi, Quoc~V Le, and Denny Zhou.
\newblock Chain-of-thought prompting elicits reasoning in large language models.
\newblock In \emph{Advances in Neural Information Processing Systems (NeurIPS)}, pages 24824--24837, 2022.

\bibitem[Xi et~al.(2025)Xi, Chen, Guo, He, Ding, Hong, Zhang, Wang, Jin, Zhou, et~al.]{xi2025rise}
Zhiheng Xi, Wenxiang Chen, Xin Guo, Wei He, Yiwen Ding, Boyang Hong, Ming Zhang, Junzhe Wang, Senjie Jin, Enyu Zhou, et~al.
\newblock The rise and potential of large language model based agents: A survey.
\newblock \emph{Science China Information Sciences}, 68\penalty0 (2):\penalty0 121101, 2025.

\bibitem[Zhang et~al.(2024{\natexlab{a}})Zhang, Yang, Hu, Wang, Li, Sun, Zhang, Zhang, Liu, Zhu, et~al.]{zhang2024proagent}
Ceyao Zhang, Kaijie Yang, Siyi Hu, Zihao Wang, Guanghe Li, Yihang Sun, Cheng Zhang, Zhaowei Zhang, Anji Liu, Song-Chun Zhu, et~al.
\newblock Proagent: building proactive cooperative agents with large language models.
\newblock In \emph{Proceedings of the AAAI Conference on Artificial Intelligence}, pages 17591--17599, 2024{\natexlab{a}}.

\bibitem[Zhang et~al.(2023{\natexlab{a}})Zhang, Du, Shan, Zhou, Du, Tenenbaum, Shu, and Gan]{zhang2023building}
Hongxin Zhang, Weihua Du, Jiaming Shan, Qinhong Zhou, Yilun Du, Joshua~B Tenenbaum, Tianmin Shu, and Chuang Gan.
\newblock Building cooperative embodied agents modularly with large language models.
\newblock In \emph{Proceedings of the International Conference on Learning Representations (ICLR)}, 2023{\natexlab{a}}.

\bibitem[Zhang et~al.(2021)Zhang, Liu, Chen, Hao, and Wang]{zhang2021deep}
Peng Zhang, Furui Liu, Zhitang Chen, Jianye Hao, and Jun Wang.
\newblock Deep reinforcement learning with causality-based intrinsic reward, 2021.
\newblock URL \url{https://openreview.net/forum?id=30I4Azqc_oP}.

\bibitem[Zhang et~al.(2023{\natexlab{b}})Zhang, Du, Huang, Wang, Wang, Fang, and Pechenizkiy]{zhang2023interpretable}
Yudi Zhang, Yali Du, Biwei Huang, Ziyan Wang, Jun Wang, Meng Fang, and Mykola Pechenizkiy.
\newblock Interpretable reward redistribution in reinforcement learning: a causal approach.
\newblock In \emph{Advances in Neural Information Processing Systems (NeurIPS)}, pages 20208--20229, 2023{\natexlab{b}}.

\bibitem[Zhang et~al.(2024{\natexlab{b}})Zhang, Du, Huang, Fang, and Pechenizkiy]{zhang2024causality}
Yudi Zhang, Yali Du, Biwei Huang, Meng Fang, and Mykola Pechenizkiy.
\newblock A causality-inspired spatial-temporal return decomposition approach for multi-agent reinforcement learning.
\newblock In \emph{NeurIPS 2024 Causal Representation Learning Workshop}, 2024{\natexlab{b}}.

\bibitem[Zhao et~al.(2023{\natexlab{a}})Zhao, Song, Yuan, Hu, Gao, Wu, Sun, and Yang]{zhao2023maximum}
Rui Zhao, Jinming Song, Yufeng Yuan, Haifeng Hu, Yang Gao, Yi~Wu, Zhongqian Sun, and Wei Yang.
\newblock Maximum entropy population-based training for zero-shot human-ai coordination.
\newblock In \emph{Proceedings of the AAAI Conference on Artificial Intelligence}, pages 6145--6153, 2023{\natexlab{a}}.

\bibitem[Zhao et~al.(2023{\natexlab{b}})Zhao, Zhou, Li, Tang, Wang, Hou, Min, Zhang, Zhang, Dong, et~al.]{zhao2023survey}
Wayne~Xin Zhao, Kun Zhou, Junyi Li, Tianyi Tang, Xiaolei Wang, Yupeng Hou, Yingqian Min, Beichen Zhang, Junjie Zhang, Zican Dong, et~al.
\newblock A survey of large language models.
\newblock \emph{arXiv preprint arXiv:2303.18223}, 2023{\natexlab{b}}.

\bibitem[Zhou et~al.(2022)Zhou, Sch{\"a}rli, Hou, Wei, Scales, Wang, Schuurmans, Cui, Bousquet, Le, et~al.]{zhou2022least}
Denny Zhou, Nathanael Sch{\"a}rli, Le~Hou, Jason Wei, Nathan Scales, Xuezhi Wang, Dale Schuurmans, Claire Cui, Olivier Bousquet, Quoc~V Le, et~al.
\newblock Least-to-most prompting enables complex reasoning in large language models.
\newblock In \emph{Proceedings of the International Conference on Learning Representations (ICLR)}, 2022.

\bibitem[Zhu et~al.(2024)Zhu, Qiao, Ou, Deng, Zhang, Lyu, Shen, Liang, Gu, and Chen]{zhu2024knowagent}
Yuqi Zhu, Shuofei Qiao, Yixin Ou, Shumin Deng, Ningyu Zhang, Shiwei Lyu, Yue Shen, Lei Liang, Jinjie Gu, and Huajun Chen.
\newblock Knowagent: knowledge-augmented planning for {LLM}-based agents.
\newblock \emph{arXiv preprint arXiv:2403.03101}, 2024.

\end{thebibliography}


\begin{thebibliography}{18}
\providecommand{\natexlab}[1]{#1}
\providecommand{\url}[1]{\texttt{#1}}
\expandafter\ifx\csname urlstyle\endcsname\relax
  \providecommand{\doi}[1]{doi: #1}\else
  \providecommand{\doi}{doi: \begingroup \urlstyle{rm}\Url}\fi

\bibitem[Carroll et~al.(2019)Carroll, Shah, Ho, Griffiths, Seshia, Abbeel, and Dragan]{carroll2019utility}
Micah Carroll, Rohin Shah, Mark~K Ho, Thomas~L Griffiths, Sanjit~A Seshia, Pieter Abbeel, and Anca Dragan.
\newblock On the utility of learning about humans for human-{AI} coordination.
\newblock In \emph{Advances in Neural Information Processing Systems (NeurIPS)}, pages 5174--5185, 2019.

\bibitem[Cohere(2024)]{cohere2024commandr}
Cohere.
\newblock The command {R} model (details and application).
\newblock \url{https://docs.cohere.com/v2/docs/command-r}, 2024.
\newblock Accessed: 2025-05-12.

\bibitem[Hoyer et~al.(2008)Hoyer, Janzing, Mooij, Peters, and Sch{\"o}lkopf]{hoyer2008nonlinear}
Patrik~O Hoyer, Dominik Janzing, Joris Mooij, Jonas Peters, and Bernhard Sch{\"o}lkopf.
\newblock Nonlinear causal discovery with additive noise models.
\newblock In \emph{Advances in Neural Information Processing Systems (NeurIPS)}, pages 689--696, 2008.

\bibitem[Jaderberg et~al.(2017)Jaderberg, Dalibard, Osindero, Czarnecki, Donahue, Razavi, Vinyals, Green, Dunning, Simonyan, Fernando, and Kavukcuoglu]{jaderberg2017population}
Max Jaderberg, Valentin Dalibard, Simon Osindero, Wojciech~M Czarnecki, Jeff Donahue, Ali Razavi, Oriol Vinyals, Tim Green, Iain Dunning, Karen Simonyan, Chrisantha Fernando, and Koray Kavukcuoglu.
\newblock Population based training of neural networks.
\newblock \emph{arXiv preprint arXiv:1711.09846}, 2017.

\bibitem[Ke et~al.(2019)Ke, Bilaniuk, Goyal, Bauer, Larochelle, Sch{\"o}lkopf, Mozer, Pal, and Bengio]{ke2019learning}
Nan~Rosemary Ke, Olexa Bilaniuk, Anirudh Goyal, Stefan Bauer, Hugo Larochelle, Bernhard Sch{\"o}lkopf, Michael~C Mozer, Chris Pal, and Yoshua Bengio.
\newblock Learning neural causal models from unknown interventions.
\newblock \emph{arXiv preprint arXiv:1910.01075}, 2019.

\bibitem[Li et~al.(2023)Li, Zhang, Sun, Du, Wen, Wang, and Pan]{li2023cooperative}
Yang Li, Shao Zhang, Jichen Sun, Yali Du, Ying Wen, Xinbing Wang, and Wei Pan.
\newblock Cooperative open-ended learning framework for zero-shot coordination.
\newblock In \emph{Proceedings of the International Conference on Machine Learning (ICML)}, pages 20470--20484, 2023.

\bibitem[Pearl(2009)]{pearl2009causality}
Judea Pearl.
\newblock \emph{Causality}.
\newblock Cambridge University Press, 2009.

\bibitem[Peng et~al.(2022)Peng, Hu, Zhang, Tang, Guo, Yi, Chen, Zhang, Du, Li, Guo, and Chen]{peng2022causality}
Shaohui Peng, Xing Hu, Rui Zhang, Ke~Tang, Jiaming Guo, Qi~Yi, Ruizhi Chen, Xishan Zhang, Zidong Du, Ling Li, Qi~Guo, and Yunji Chen.
\newblock Causality-driven hierarchical structure discovery for reinforcement learning.
\newblock In \emph{Advances in Neural Information Processing Systems (NeurIPS)}, pages 20064--20076, 2022.

\bibitem[Peters et~al.(2017)Peters, Janzing, and Schlkopf]{peters2017elements}
Jonas Peters, Dominik Janzing, and Bernhard Schlkopf.
\newblock \emph{Elements of Causal Inference: Foundations and Learning Algorithms}.
\newblock The MIT Press, 2017.

\bibitem[Qiao et~al.(2024)Qiao, Fang, Zhang, Zhu, Chen, Deng, Jiang, Xie, Huang, and Chen]{qiao2024agent}
Shuofei Qiao, Runnan Fang, Ningyu Zhang, Yuqi Zhu, Xiang Chen, Shumin Deng, Yong Jiang, Pengjun Xie, Fei Huang, and Huajun Chen.
\newblock Agent planning with world knowledge model.
\newblock In \emph{Advances in Neural Information Processing Systems (NeurIPS)}, volume~37, pages 114843--114871, 2024.

\bibitem[Seitzer et~al.(2021)Seitzer, Sch{\"o}lkopf, and Martius]{seitzer2021causal}
Maximilian Seitzer, Bernhard Sch{\"o}lkopf, and Georg Martius.
\newblock Causal influence detection for improving efficiency in reinforcement learning.
\newblock In \emph{Advances in Neural Information Processing Systems (NeurIPS)}, pages 22905--22918, 2021.

\bibitem[Spirtes et~al.(2000)Spirtes, Glymour, and Scheines]{spirtescausation}
Peter Spirtes, Clark Glymour, and Richard Scheines.
\newblock \emph{Causation, Prediction, and Search}.
\newblock The MIT Press, 2000.

\bibitem[Strouse et~al.(2021)Strouse, McKee, Botvinick, Hughes, and Everett]{strouse2021collaborating}
DJ~Strouse, Kevin~R McKee, Matt Botvinick, Edward Hughes, and Richard Everett.
\newblock Collaborating with humans without human data.
\newblock In \emph{Advances in Neural Information Processing Systems (NeurIPS)}, pages 14502--14515, 2021.

\bibitem[Tesauro(1994)]{tesauro1994td}
Gerald Tesauro.
\newblock {TD}-{G}ammon, a self-teaching backgammon program, achieves master-level play.
\newblock \emph{Neural Computation}, 6\penalty0 (2):\penalty0 215--219, 1994.

\bibitem[Wang et~al.(2022)Wang, Wei, Schuurmans, Le, Chi, Narang, Chowdhery, and Zhou]{wang2022self}
Xuezhi Wang, Jason Wei, Dale Schuurmans, Quoc~V Le, Ed~H Chi, Sharan Narang, Aakanksha Chowdhery, and Denny Zhou.
\newblock Self-consistency improves chain of thought reasoning in language models.
\newblock In \emph{Proceedings of the International Conference on Learning Representations (ICLR)}, 2022.

\bibitem[Williams and Rasmussen(2006)]{williams2006gaussian}
Christopher~KI Williams and Carl~Edward Rasmussen.
\newblock \emph{Gaussian {P}rocesses for {M}achine {L}earning}, volume~2.
\newblock MIT Press Cambridge, MA, 2006.

\bibitem[Zhang et~al.(2024)Zhang, Yang, Hu, Wang, Li, Sun, Zhang, Zhang, Liu, Zhu, et~al.]{zhang2024proagent}
Ceyao Zhang, Kaijie Yang, Siyi Hu, Zihao Wang, Guanghe Li, Yihang Sun, Cheng Zhang, Zhaowei Zhang, Anji Liu, Song-Chun Zhu, et~al.
\newblock Proagent: building proactive cooperative agents with large language models.
\newblock In \emph{Proceedings of the AAAI Conference on Artificial Intelligence}, pages 17591--17599, 2024.

\bibitem[Zhao et~al.(2023)Zhao, Song, Yuan, Hu, Gao, Wu, Sun, and Yang]{zhao2023maximum}
Rui Zhao, Jinming Song, Yufeng Yuan, Haifeng Hu, Yang Gao, Yi~Wu, Zhongqian Sun, and Wei Yang.
\newblock Maximum entropy population-based training for zero-shot human-ai coordination.
\newblock In \emph{Proceedings of the AAAI Conference on Artificial Intelligence}, pages 6145--6153, 2023.

\end{thebibliography}
